\documentclass{article}

\usepackage[final, nonatbib]{neurips_2024}

\usepackage[utf8]{inputenc} % allow utf-8 input
\usepackage[T1]{fontenc}    % use 8-bit T1 fonts
\usepackage{hyperref}       % hyperlinks
\usepackage{url}            % simple URL typesetting
\usepackage{booktabs}       % professional-quality tables
\usepackage{amsfonts}       % blackboard math symbols
\usepackage{nicefrac}       % compact symbols for 1/2, etc.
\usepackage{microtype}      % microtypography
\usepackage{wrapfig}

\usepackage{subfigure}
\usepackage[ruled]{algorithm2e}
\usepackage{bbm}
\usepackage{amsfonts}
\usepackage{epsfig}
\usepackage{graphicx}
\usepackage{amsmath,bm}
\usepackage{amssymb}
\usepackage{bbding}
\usepackage{color}
\usepackage{xcolor}
\usepackage{cleveref}

\usepackage{wrapfig}

\crefname{figure}{figure}{figures}
\Crefname{figure}{Figure}{Figures}
\hypersetup{colorlinks,linkcolor={red},citecolor={green},urlcolor={blue}}

\hypersetup{
    colorlinks=true,
    urlcolor=magenta,
}

\def\ie{i.e.,~}               % that is, in other words
\title{4Diffusion: Multi-view Video Diffusion Model for 4D Generation}
\author{%
  Haiyu Zhang\textsuperscript{1,2}\thanks{Work done when Haiyu Zhang interned at Shanghai AI Laboratory.},
  Xinyuan Chen\textsuperscript{2},
  Yaohui Wang\textsuperscript{2},
  Xihui Liu\textsuperscript{3},
  \textbf{Yunhong Wang\textsuperscript{1}},
  \textbf{Yu Qiao\textsuperscript{2}\thanks{Corresponding author}}\\
    {\textsuperscript{1}Beihang University \ \ 
    \textsuperscript{2}Shanghai AI Laboratory \ \ 
    \textsuperscript{3}The University of Hong Kong
    }\\
    \textsuperscript{1}\texttt{\{zhyzhy,yhwang\}@buaa.edu.cn} \ \ 
 \textsuperscript{2}\texttt{\{chenxinyuan,wangyaohui,qiaoyu\}@pjlab.org.cn}\\
  \textsuperscript{3}\texttt{xihuiliu@eee.hku.hk}\\
    \url{https://aejion.github.io/4diffusion} \\
}

\begin{document}

\maketitle

\begin{figure}[htb]
    \centering
    \vspace{-10mm}
    \includegraphics[width=\linewidth]{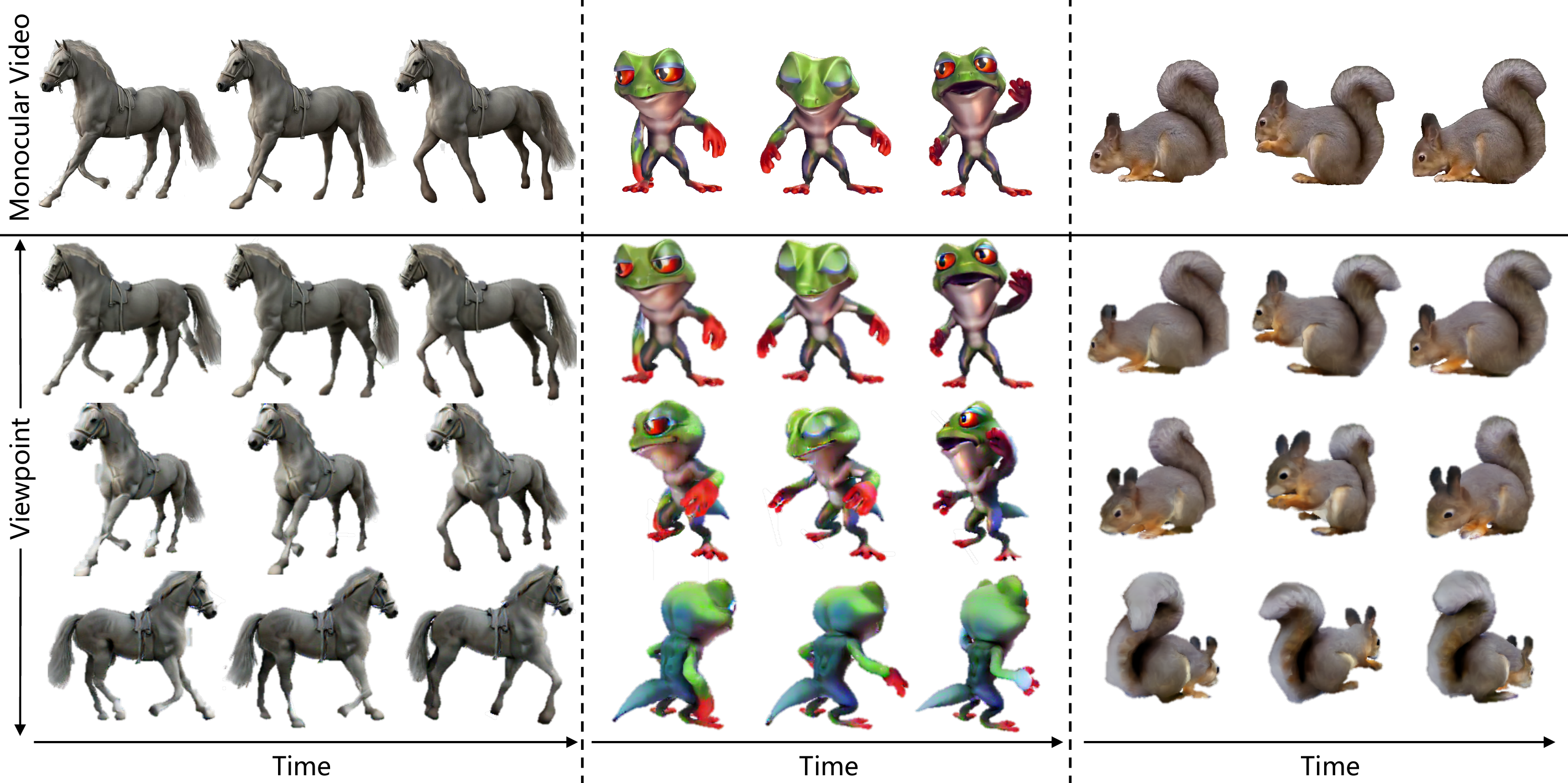}
    \caption{\textbf{4Diffusion} generates spatial-temporally consistent 4D contents from monocular videos.}
    \label{fig:teasers}
\end{figure}

\begin{abstract}
  Current 4D generation methods have achieved noteworthy efficacy with the aid of advanced diffusion generative models. However, these methods lack multi-view spatial-temporal modeling and encounter challenges in integrating diverse prior knowledge from multiple diffusion models, resulting in inconsistent temporal appearance and flickers. In this paper, we propose a novel 4D generation pipeline, namely \textbf{4Diffusion}, aimed at generating spatial-temporally consistent 4D content from a monocular video. We first design a unified diffusion model tailored for multi-view video generation by incorporating a learnable motion module into a frozen 3D-aware diffusion model to capture multi-view spatial-temporal correlations. After training on a curated dataset, our diffusion model acquires reasonable temporal consistency and inherently preserves the generalizability and spatial consistency of the 3D-aware diffusion model. Subsequently, we propose 4D-aware Score Distillation Sampling loss, which is based on our multi-view video diffusion model, to optimize 4D representation parameterized by dynamic NeRF. This aims to eliminate discrepancies arising from multiple diffusion models, allowing for generating spatial-temporally consistent 4D content. Moreover, we devise an anchor loss to enhance the appearance details and facilitate the learning of dynamic NeRF. Extensive qualitative and quantitative experiments demonstrate that our method achieves superior performance compared to previous methods.
\end{abstract}

\section{Introduction}
\label{sec:intro}
In recent years, diffusion models have significantly impacted the era of image, video, and 3D generation. With the support of large-scale text-to-image diffusion models \cite{rombach2022high, deepfloyd} and 3D-aware diffusion models \cite{shi2023mvdream, liu2023zero, wang2023imagedream}, many works \cite{lin2023magic3d, qian2023magic123, tang2023make, chen2023fantasia3d, melas2023realfusion, raj2023dreambooth3d, tang2023dreamgaussian, liu2023humangaussian, wang2024prolificdreamer,lin2023consistent123} leverage Score Distillation Sampling (SDS) \cite{poole2022dreamfusion} to distill the prior knowledge from diffusion models to optimize a 3D shape parameterized by NeRF \cite{mildenhall2021nerf} or 3DGS \cite{kerbl20233d}. Although they have attained faithful results, they only focus on creating static 3D shapes, neglecting the dynamics of objects in the real world.

Generating 4D content, \ie dynamic 3D content, holds diverse applications in the virtual realm, including digital human, gaming, media, and AR/VR. The main challenge lies in creating 4D content with vivid motion and high-quality spatial-temporal consistency. The pioneering study MAV3D \cite{singer2023text} introduces a two-stage method, which first learns a static 3D shape with a text-to-image diffusion model and then deforms the static 3D shape with a text-to-video diffusion model \cite{singer2022make}. However, MAV3D encounters the Janus problem and generates 4D contents with poor appearance and motion \cite{bahmani20234d}. To overcome these issues, the following works \cite{bahmani20234d, zheng2023unified, zhao2023animate124, ling2023align} employ multiple diffusion models for distinct purposes. Specifically, these methods leverage 3D-aware diffusion models \cite{shi2023mvdream, liu2023zero} and text-to-image diffusion models \cite{rombach2022high} to achieve spatial consistency and visually appealing appearance. Akin to MAV3D, they utilize video diffusion models \cite{zeroscope, wang2023videofactory, singer2022make} to add motion to create 4D content.

\begin{figure}[htb]
    \centering
    \includegraphics[width=\linewidth]{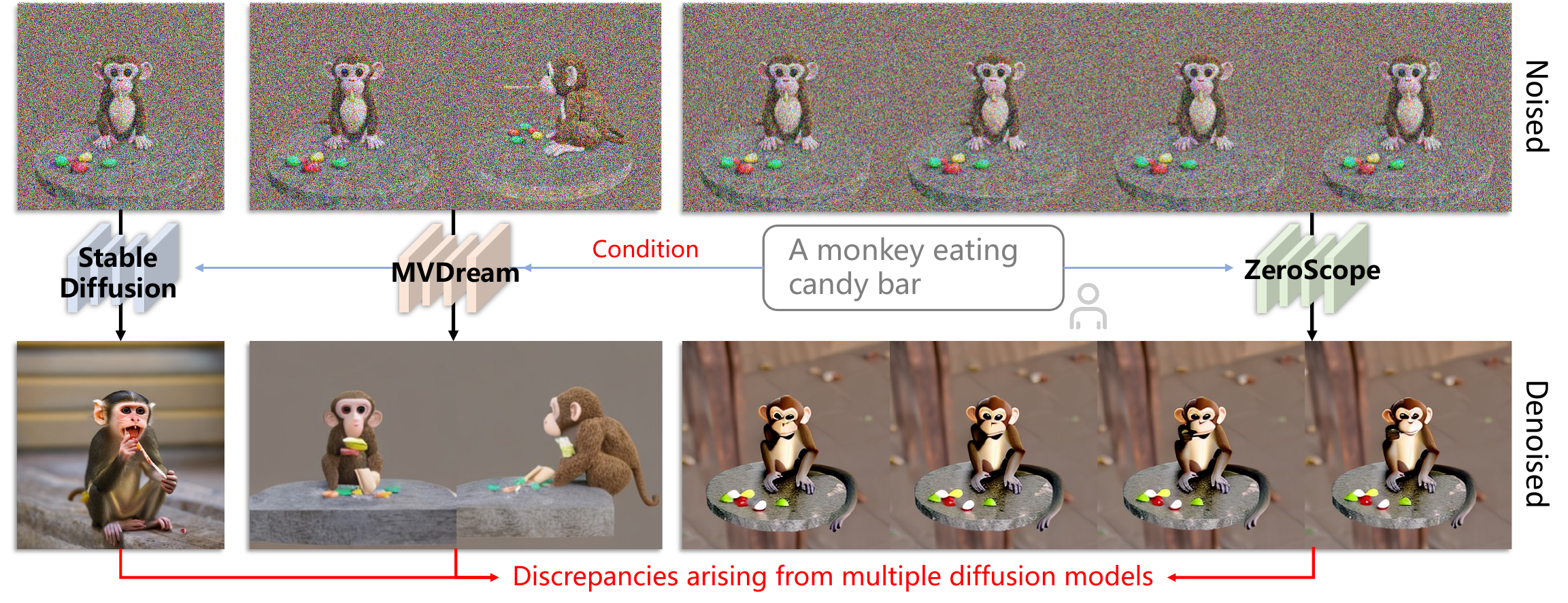}
    \vspace{-1.5em}
    \caption{\textbf{Challenges.} The denoised images from Stable Diffusion (SD) \cite{rombach2022high}, MVDream \cite{shi2023mvdream}, and ZeroScope \cite{zeroscope}. These diffusion models can not provide multi-view spatial-temporal guidance and exhibit discrepancies, making their integration challenging.
    }
    \label{fig:teaser}
\end{figure}

The aforementioned methods utilize multiple diffusion models for 4D generation. As Fig.~\ref{fig:teaser} illustrates, when diffusing images rendered from a 3D model, the 3D-aware diffusion model \cite{shi2023mvdream} generates multi-view images to address the spatial ambiguity. On the other hand, the 2D image diffusion model \cite{rombach2022high} produces a clean image with subtle details to refine appearance. The 2D video diffusion model \cite{zeroscope} generates dynamic frames to ensure temporal consistency within the same viewpoint. However, there is no accurate guidance to ensure multi-view spatial-temporal consistency due to the lack of multi-view spatial-temporal modeling. Moreover, it is challenging to integrate diverse prior knowledge from multiple diffusion models, often leading to inconsistent temporal appearance and flickers as shown in the second row of Fig.~\ref{fig:4d_generation}.

In this paper, we present a novel 4D generation pipeline, namely \textbf{4Diffusion}, to create high-quality spatial-temporally consistent 4D content from a monocular video. Specifically, we propose a unified diffusion model, 4DM, to capture multi-view spatial-temporal correlations for multi-view video generation. To achieve this, we construct 4DM based on the powerful pre-trained 3D-aware diffusion model \cite{wang2023imagedream}, which already ensures high-quality multi-view spatial consistency. We then seamlessly integrate a motion module into the 3D-aware diffusion model to extend the temporal modeling capability. Contrary to previous attempts \cite{guo2023animatediff, guo2023i2v} that typically demand extensive large-scale video datasets for tuning the motion module, 4DM achieves reasonable temporal consistency and captures multi-view spatial-temporal correlations after training on only hundreds of multi-view videos. Importantly, we keep the parameters of the 3D-aware diffusion model unchanged to preserve the generalization ability and spatial consistency. 4DM provides multi-view spatial-temporal guidance for 4D generation. Therefore, we propose 4D-aware SDS loss to distill prior knowledge from 4DM to optimize 4D content parameterized by dynamic NeRF. This approach eliminates discrepancies arising from multiple diffusion models and stabilizes the optimizing process. Moreover, we use 4DM to generate anchor videos conditioned on the input monocular video and devise an anchor loss to enhance the appearance details, facilitating the learning of dynamic NeRF. Finally, we generate 4D content with high-quality spatial-temporal consistency and vibrant motion coherence with the input video as shown in Fig.~\ref{fig:teasers}. Qualitative and quantitative experiments demonstrate that our method achieves state-of-the-art performance on multi-view video generation and 4D generation from monocular videos.

To summarize, our contributions are as follows: \textbf{1)} We present \textbf{4Diffusion}, a novel 4D generation pipeline that generates high-quality spatial-temporal consistent 4D content from a monocular video with a multi-view video diffusion model. \textbf{2)} We propose a multi-view video diffusion model, 4DM, which provides multi-view spatial-temporal guidance for 4D generation. It trains on only hundreds of curated high-quality multi-view videos to capture multi-view spatial-temporal correlations. \textbf{3)} We combine 4D-aware SDS loss and an anchor loss based on 4DM to optimize dynamic NeRF, which stabilizes the training process and allows for generating high-quality 4D content. 

\section{Related Work}

Recent breakthroughs in multiple research domains have significantly accelerated progress in 4D generation task. Here, we discuss the most relevant fields, including 3D generation, video and 3D-aware diffusion models, and 4D generation.

\noindent\textbf{3D Generation.}  Recent studies in 3D generation can be classified into three categories: 3D generative methods \cite{wang2023rodin, gupta20233dgen, shue20233d, muller2023diffrf, yariv2023mosaic, anciukevivcius2023renderdiffusion, xu2023dmv3d, erkocc2023hyperdiffusion}, feed forward methods \cite{hong2023lrm, tang2024lgm, li2023instant3d, zou2023triplane}, and diffusion prior-based methods \cite{lin2023magic3d, qian2023magic123, tang2023make, chen2023fantasia3d, melas2023realfusion, raj2023dreambooth3d, tang2023dreamgaussian, liu2023humangaussian, wang2024prolificdreamer,lin2023consistent123}. Inspired by the advancements in 2D content creation, 3D generative methods utilize the robust diffusion \cite{wang2023rodin} or flow-based \cite{yariv2023mosaic} backbone to generate 3D data represented by Signed Distance Function (SDF) \cite{yariv2023mosaic}, voxel grid \cite{muller2023diffrf}, triplane \cite{chan2022efficient, gupta20233dgen, shue20233d}, or weights of neural network \cite{erkocc2023hyperdiffusion}. However, these methods require time-consuming pre-training to fit each 3D data and are limited to creating a single category. Feed forward methods \cite{hong2023lrm, li2023instant3d} adopt image features extracted from the pre-trained visual encoder DINO \cite{caron2021emerging} to reconstruct 3D representations through a highly scalable and efficient transformer-based decoder. Although they can produce a 3D shape in a few seconds, they demand extensive training on large-scale 3D datasets, which is impractical with limited 4D datasets for 4D generation. Furthermore, diffusion prior-based methods distill prior knowledge from diffusion generative models via SDS \cite{poole2022dreamfusion} to optimize 3D representations, enabling the generation of high-quality 3D shapes with strong generalizability. In contrast to static 3D generation, our method focuses on creating 4D content.

\noindent\textbf{Video and 3D-aware Diffusion Models.} With the success of large-scale text-to-image diffusion models \cite{rombach2022high,deepfloyd}, recent works attempt to use diffusion models to generate more complex signals, including video and 3D. AnimateDiff \cite{guo2023animatediff} inserts a learnable motion module into the frozen text-to-image model for video generation, which preserves the efficacy of the text-to-image model while successfully modeling temporal information. Recent 3D-aware diffusion model Zero-1-to-3 \cite{liu2023zero} adopts a stable diffusion model conditioned on relative camera pose and a single image for novel view synthesis. However, this method still suffers from the Janus problem and content drafting problem \cite{shi2023mvdream} due to the lack of explicit 3D modeling. Approaches like \cite{liu2023syncdreamer, shi2023mvdream, yang2023consistnet, long2023wonder3d, wang2023imagedream} leverage 3D-aware attention block to model the joint probability distribution of multi-view images, leading to spatially consistent generation. However, these approaches are incapable of producing multi-view consistent videos, due to the absence of temporal or spatial modeling.

\begin{figure}[t]
    %\small\includegraphics[width=0.8\linewidth]{assets/dataset.pdf}
    \includegraphics[width=\linewidth]{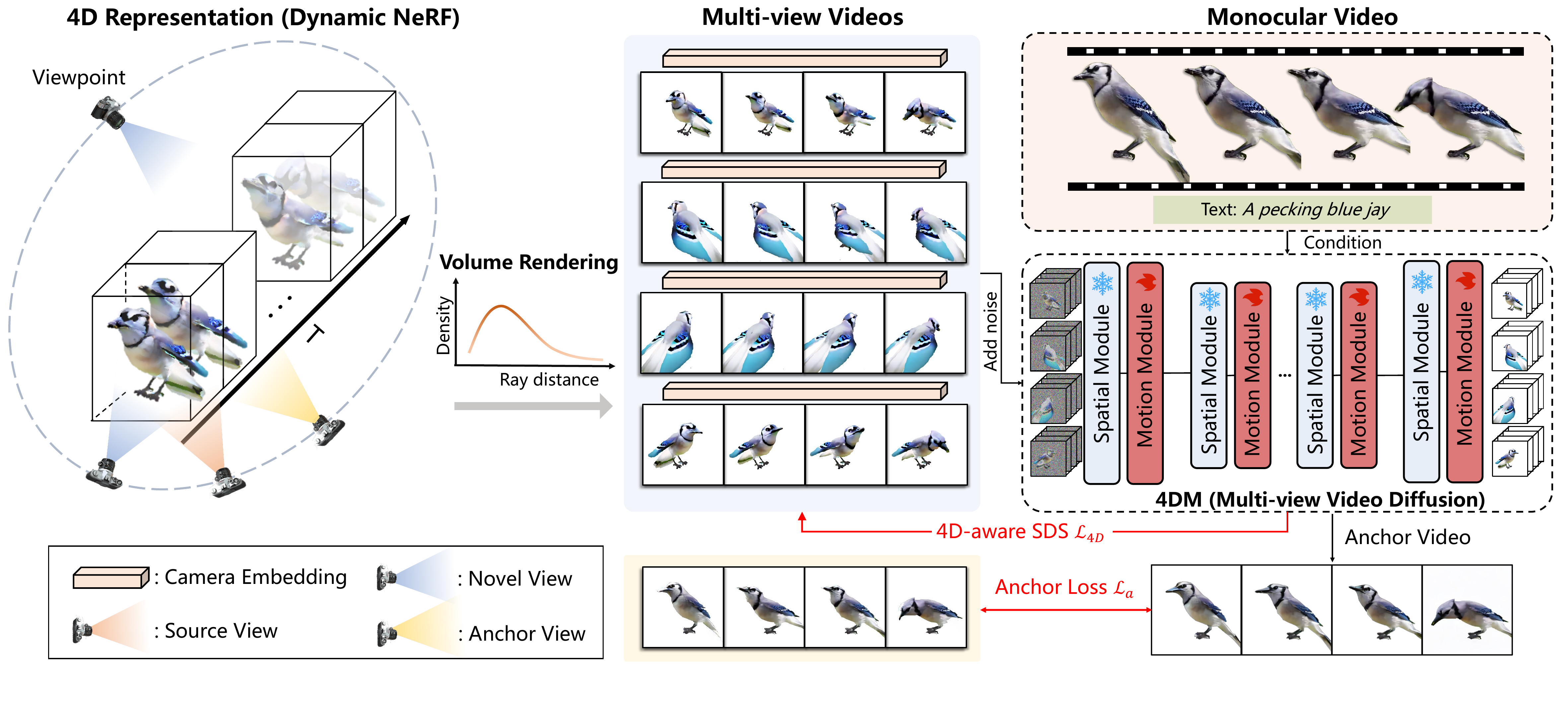}
    \vspace{-2em}
    \caption{\textbf{4Diffusion overview.} Our method first trains a unified diffusion, named 4DM, by inserting a learnable motion module at the end of each frozen spatial module of ImageDream to capture multi-view spatial-temporal correlations. Given a monocular video and text prompt, 4DM can produce consistent multi-view videos. Then, we combine 4D-aware SDS and an anchor loss based on 4DM to optimize 4D content parameterized by Dynamic NeRF.}
    \label{fig:pipeline}
\end{figure}

\textbf{4D Generation.} Recently, several works have delved into 4D generation from various user-friendly prompts, such as text \cite{singer2023text, bahmani20234d, zheng2023unified, ling2023align}, a single image \cite{zhao2023animate124, zheng2023unified}, and a monocular video \cite{ren2023dreamgaussian4d, jiang2023consistent4d, yin20234dgen}. The pioneering study MAV3D \cite{singer2023text} proposes a two-stage method to optimize 4D representation, \ie Hexplane \cite{cao2023hexplane}, with both text-to-image and text-to-video diffusion models in a static-to-dynamic manner. To generate 4D contents with realistic appearance, Dream-in-4D \cite{zheng2023unified} and 4D-fy \cite{bahmani20234d} combine hybrid diffusion models. Specifically, they utilize 3D-aware and 2D diffusion guidance to learn a static 3D representation and incorporate video diffusion guidance to add motion. However, these diffusion models can not offer multi-view spatial-temporally consistent guidance and it is difficult to integrate diverse prior knowledge from multiple diffusion models, resulting in suboptimal results. In contrast to these approaches, we design a unified model to capture multi-view spatial-temporal correlations for 4D generation.

Similar to us, \cite{jiang2023consistent4d, ren2023dreamgaussian4d, zeng2024stag4d, yin20234dgen, wu2024sc4d} generate 4D content from a monocular video. Consistent4D \cite{jiang2023consistent4d} introduces an interpolation-driven loss between two adjacent frames to enhance spatial-temporal consistency. However, Consistent4D lacks temporal modeling cross frames. DreamGaussian4D \cite{ren2023dreamgaussian4d}, 4DGen \cite{yin20234dgen}, and SC4D \cite{wu2024sc4d} combine 4D Gaussian Splatting \cite{wu20234d, huang2023sc} into 4D generation pipeline. Although they notably reduce optimization time, they may result in blurred appearance and inaccurate geometry due to the explicit characteristics of Gaussians. STAG4D \cite{zeng2024stag4d} proposes a training-free strategy to generate sparse anchor multi-view videos for 4D generation. In contrast, we propose a multi-view video diffusion model to provide multi-view spatial-temporal consistency guidance for 4D generation.

\section{Method}

Given a monocular video $V=\{I_{j}|j=1,2,...,T\}$ with $T$ frames and an optional textual caption, our goal is to generate a high-quality spatial-temporally consistent 4D content, capable of rendering from any novel viewpoint across the temporal dimension. In Sec.~\ref{sec:preliminary}, we talk about 3D-aware diffusion models, employed as the initialization of our unified diffusion model. In Sec.~\ref{sec:video_diffusion}, we propose a unified diffusion model 4DM to capture multi-view spatial-temporal correlations for multi-view video generation. Subsequently, we elaborate on distilling prior knowledge from 4DM to optimize 4D content parameterized by dynamic NeRF and devise an anchor loss to enhance the appearance details, as detailed in Sec.~\ref{sec:optimize}. Fig.~\ref{fig:pipeline} shows the overall pipeline of our method.

% As shown in Fig.~\ref{fig:pipeline}, we propose a unified diffusion model 4DM to capture multi-view spatial-temporal correlations for multi-view video generation, which is achieved by inserting a learnable motion module into a frozen 3D-aware diffusion model, as described in Sec.~\ref{sec:video_diffusion}. Subsequently, we elaborate on distilling prior knowledge from 4DM to optimize 4D representation parameterized by dynamic NeRF and devise an anchor loss to enhance the appearance details, as detailed in Sec.~\ref{sec:optimize}.

\subsection{Preliminary: 3D-aware Diffusion Models}
\label{sec:preliminary}
3D-aware diffusion models learn spatial relationships from multi-view images for 3D generation and can serve as an initialization of our unified diffusion model. Recent works \cite{liu2023syncdreamer, yang2023consistnet, long2023wonder3d} mainly focus on generating multi-view images from predetermined sparse viewpoints and necessitate additional algorithms for 3D reconstruction. Although we can extend these methods to generate multi-view videos and employ 4D reconstruction algorithms, it is challenging to reconstruct high-quality 4D content from a limited number of viewpoints. Therefore, we design our unified diffusion model to generate multi-view videos from arbitrary viewpoints and choose ImageDream \cite{wang2023imagedream} as initialization. Given four arbitrary orthogonal viewpoints under canonical coordination and a single image with an optional textual caption, ImageDream can synthesize four multi-view images that align coherently with the input. Specifically, ImageDream utilizes an adapter similar to IP-Adapter \cite{ye2023ip} to inject image prompts and a 3D self-attention module to capture spatial relationships.

\subsection{4DM: Multi-view Video Diffusion Model}
\label{sec:video_diffusion}

To maintain the spatial consistency and mitigate training complexity, we design our multi-view video diffusion model 4DM based on a pre-trained 3D-aware diffusion model (\ie ImageDream \cite{wang2023imagedream}). Given a monocular video with an optional text prompt and four orthogonal novel viewpoints under canonical coordination, 4DM aims to generate four spatial-temporally consistent videos.

\begin{figure}[t]
    \centering
    \begin{minipage}[b]{0.45\textwidth}
        \centering
        \includegraphics[width=\textwidth]{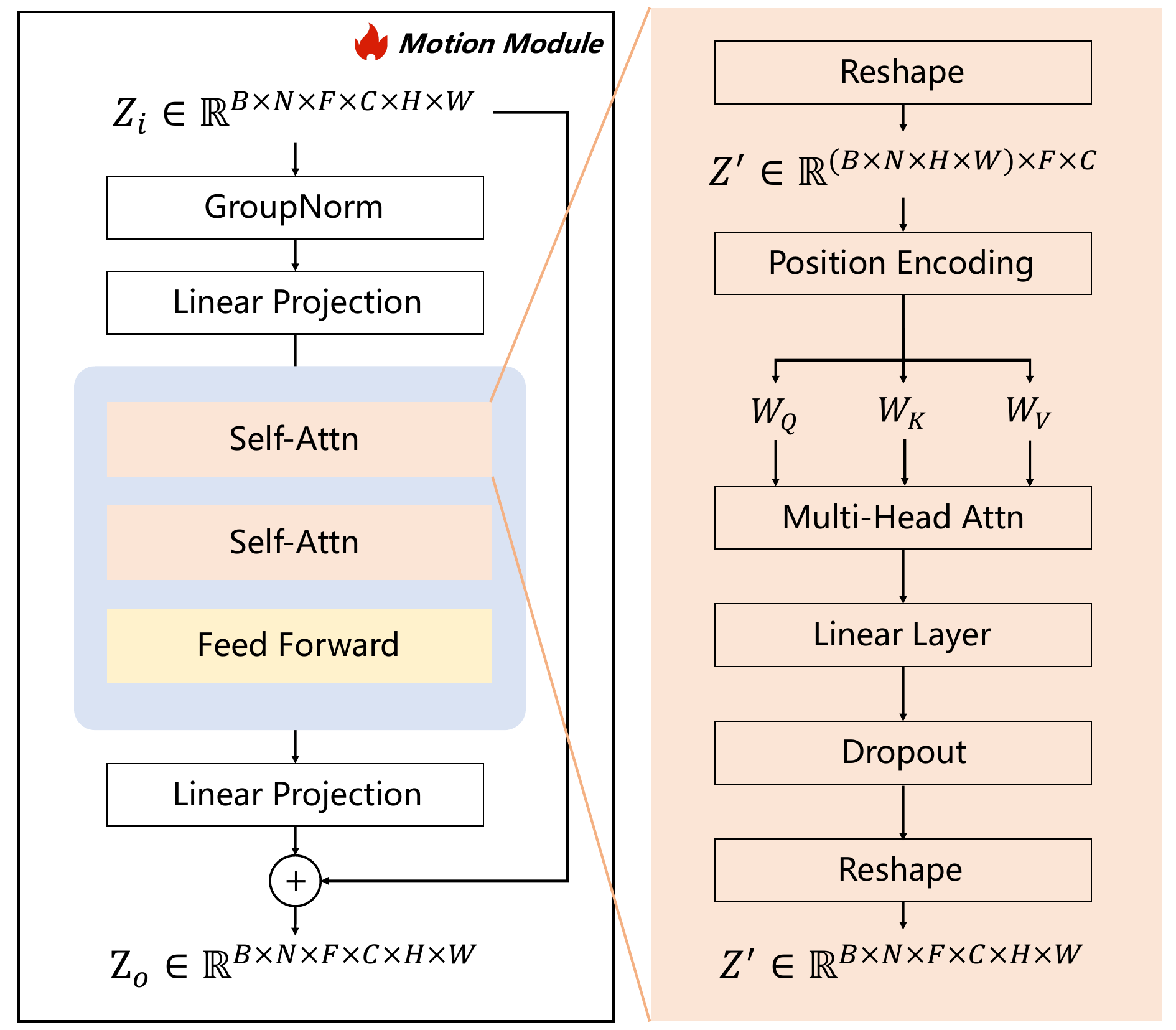}
        \caption{The detailed overview of the architecture of motion module.}
        \label{fig:motion_module}
    \end{minipage}
    \hspace{0.05\textwidth}
    \begin{minipage}[b]{0.45\textwidth}
        \centering
        \includegraphics[width=\textwidth]{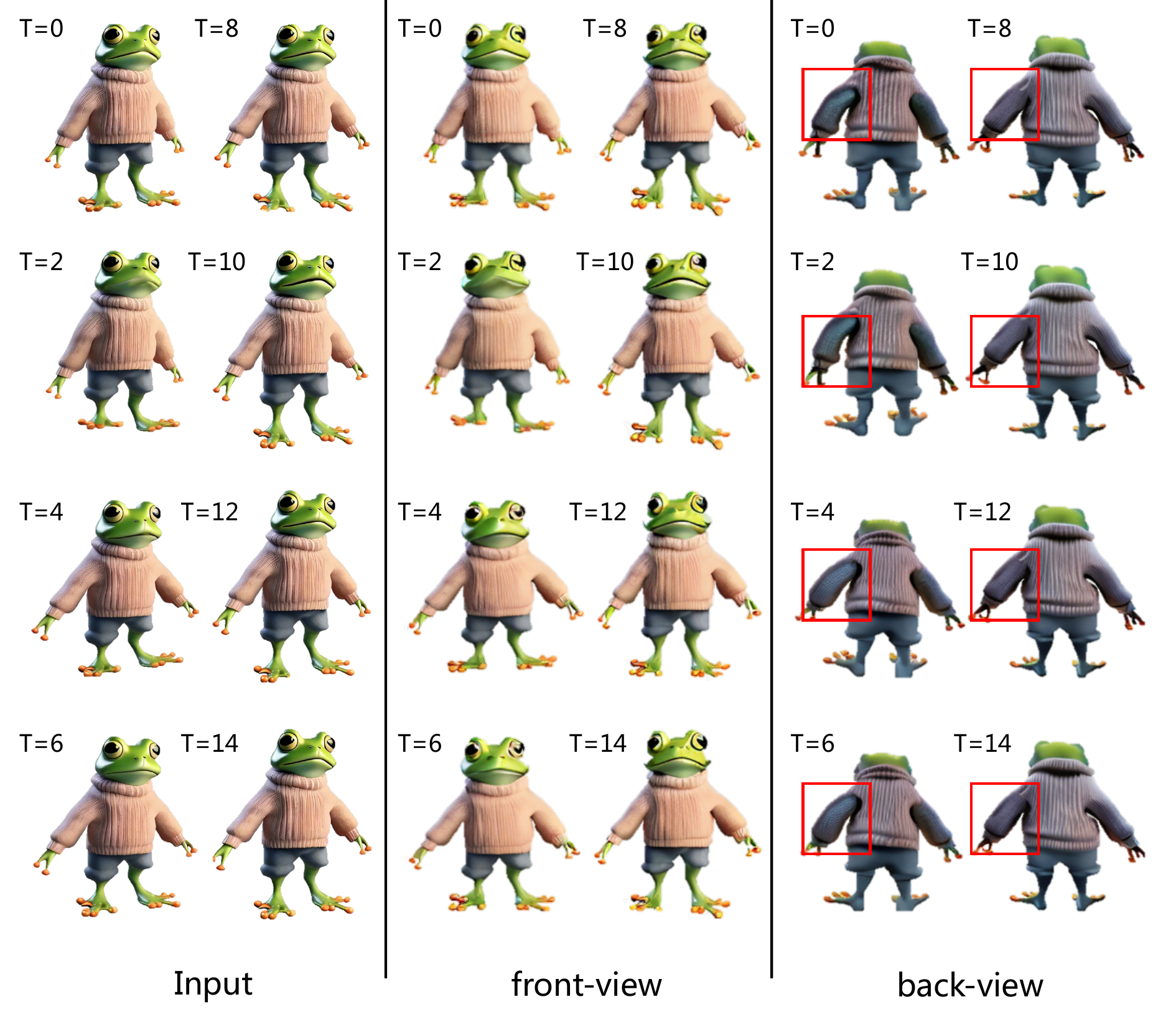}
        \caption{The illustration of multi-view video generation when input video exceeds 8 frames.}
        \label{fig:multi-4d}
    \end{minipage}
\end{figure}

Although we can directly use the original ImageDream to generate a set of individual multi-view images to form multi-view videos, the result lacks temporal consistency as ImageDream has no layer for temporal modeling, as shown in Fig.~\ref{fig:multi-view-video}. We thus add a zero-initialized motion module at the end of each block of the UViT network of ImageDream. Specifically, each motion module begins with group normalization and a linear projection, followed by two self-attention blocks and one feed-forward block. A final linear projection is then applied, after which the residual hidden feature is added back at the end of each motion module as detailed in Fig.~\ref{fig:motion_module}. Then, each attention block $i$ of 4DM includes a spatial module and a motion module $l_{m}^{i}$. The spatial module comprises a 3D self-attention module $l_{s}^{i}$ and a cross-attention module. We first concatenate the monocular video latent and four multi-view video latents encoded by VAE \cite{kingma2013auto} to obtain a batch $B$ of latents $\mathbf{Z} \in \mathbb{R}^{B\times F\times N \times C \times H \times W}$, where $C$ is the number of channels, $H$ and $W$ are spatial resolutions, $N=5$ is the number of viewpoints, and $F$ is the number of frames. Subsequently, we reshape the temporal axis into the batch dimension and independently process multi-view video latents through the 3D self-attention module,
\begin{align}
    & \mathbf{Z_{s}} \longleftarrow \textbf{Reshape}(\textbf{Z}, B\ F\ N\ C\ H\ W \rightarrow (B\ F)\ N\ H\ W\ C), \\
    & \mathbf{Z_{s}} \longleftarrow l_{s}^{i}(\mathbf{Z_{s}}), \\
    & \mathbf{Z_{s}} \longleftarrow \textbf{Reshape}(\mathbf{Z_{s}}, (B\ F)\ N\ H\ W\ C \rightarrow B\ F\ N\ C\ H\ W).
\end{align}

Then, we use the adapter in ImageDream to individually process the input video frames and output the video features to perform cross-attention operations. Here, we also reshape the temporal axis into the batch dimension to prevent dimensional confusion. Furthermore, for the motion module, we perform self-attention exclusively along the temporal axis by reshaping the spatial dimensions and the viewpoint dimension into the batch dimension,
\begin{align}
    & \mathbf{Z'} \longleftarrow \textbf{Reshape}(\textbf{Z}, B\ F\ N\ C\ H\ W \rightarrow (B\ N\ H\ W)\ F\ C), \\
    & \mathbf{Z'} \longleftarrow l_{m}^{i}(\mathbf{Z'}), \\
    & \mathbf{Z'} \longleftarrow \textbf{Reshape}(\mathbf{Z'}, (B\ N\ H\ W)\ F\ C \rightarrow B\ F\ N\ C\ H\ W).
\end{align}

We utilize Objaverse dataset \cite{deitke2023objaverse} to train 4DM. Although Objaverse provides nearly 44K animated 3D shapes, rendering multi-view videos and training a diffusion model are time- and computation-consuming using the entire dataset. Moreover, it is worth noting that the Objaverse dataset contains a significant amount of flawed data. Consequently, we manually select a curated subset of 926 high-quality animated 3D shapes from Objaverse dataset \cite{deitke2023objaverse}. We render multi-view videos from those animated 3D shapes to tune our motion module while holding the parameters of the origin ImageDream frozen. Surprisingly, 4DM successfully learns reasonable temporal dynamics and preserves the characteristics of the origin ImageDream model, including generalization ability, spatial consistency, and image understanding ability, even when trained on a small curated dataset. As Fig.~\ref{fig:multi-view-video} illustrates, 4DM generates multi-view spatial-temporal consistent videos, surpassing the performance of ImageDream. For more details on our dataset, please refer to supplementary material.

\noindent\textbf{Training Objectives.} For each animated 3D shape from our dataset, we render a monocular video $V_{m}$ with a random viewpoint and four videos $V_{o}$ with orthogonal viewpoints $\mathbf{c}_{\mathit{mv}}^{\mathit{v}}$ and select $F = 8$ frames from each video at a stride of 4 to create our multi-view video dataset $\cal{X}_{\mathit{mv}}^{\mathit{v}}=\{\mathbf{x}_{\mathit{mv}}^{\mathit{v}}, \mathit{y}, \mathbf{x}_{\mathit{r}}^{\mathit{v}}, \mathbf{c}_{\mathit{mv}}^{\mathit{v}}\}$. Here, $\mathbf{x}_{\mathit{r}}^{\mathit{v}}$ and $\mathbf{x}_{\mathit{mv}}^{\mathit{v}}$ represent the video clips from $V_{m}$ and $V_{o}$. $\mathit{y}$ is the text prompt captioned by Cap3D \cite{luo2023scalable}. Then, we use $\cal{X}_{\mathit{mv}}^{\mathit{v}}$ following the diffusion loss to train 4DM,
\begin{equation}
\begin{split}
    \cal L_{\mathit{MV}}(\theta, \cal X_{\mathit{mv}}) &= \mathbb{E}_{\mathbf{x}, y, \mathbf{x}_{r}, \mathbf{c}, t, \epsilon}\left[ \Vert \epsilon - \epsilon_\theta(\mathbf{x}^p; y, \mathbf{x}_{r}^{p}, \mathbf{c}^p,t) \Vert_{2}^{2} \right], \label{eq:diffusion_loss} \\
    \text{where,~} (\mathbf{x}^p, \mathbf{x}_{r}^{p}, \mathbf{c}^p) &= \begin{cases} 
    (\mathbf{x}_{\mathit{mv}}, \textbf{0}, \textbf{0}), & \text{with probability } p  \\
    (\mathbf{x}_{\mathit{mv}}, \mathbf{x}_{\mathit{r}}^{\mathit{v}}, \mathbf{c}_{\mathit{mv}}^{\mathit{v}}), & \text{with probability } 1-p 
    \end{cases}
\end{split}
\end{equation}
here, $\mathbf{x}_{\mathit{mv}}$ represent the noisy video latents derived from $\mathbf{x}_{\mathit{mv}}^{\mathit{v}}$. These latents are initially encoded by VAE and subsequently noised by random noise $\epsilon$ at a diffusion timestep $t$. For more details about the noising process, please refer to \cite{rombach2022high}. $\epsilon_\theta$ is 4DM model parametrized by $\theta$. During the training of 4DM, we ensure that our training data does not overlap with the test data used in our experiments.

\begin{figure}[t]
    \centering
    %\small\includegraphics[width=0.8\linewidth]{assets/teaser.pdf}
    \includegraphics[width=\linewidth]{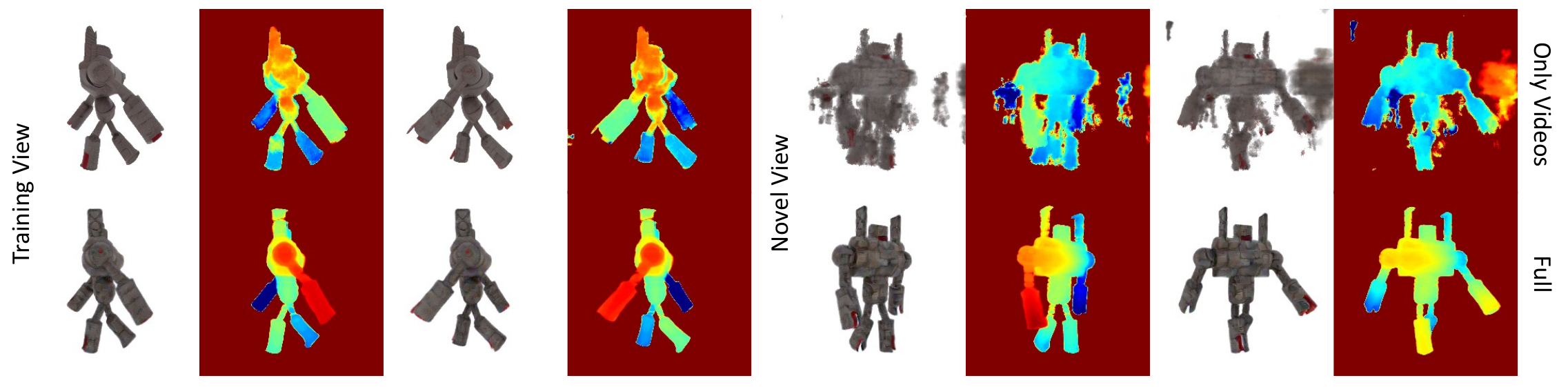}
    \vspace{-1.5em}
    \caption{Illustration of directly optimizing on the generated multi-view videos.}
    \label{fig:direct_on_videos}
\end{figure}

\subsection{4D Generation}
\label{sec:optimize}

\textbf{Dynamic NeRF Representation.} Recent methods \cite{li2022neural, fridovich2023k} use neural networks or explicit spatial grids to map a 6D spatial-temporal coordinate $(x,\mathrm{d},\mathbf{t})$ to density $\mathbf{\tau}(x,\mathbf{t}) \in \mathbb{R}_{+}$ and view-dependent color $c(x,\mathrm{d},\mathbf{t}) \in \mathbb{R}^{3}_{+}$ of dynamic scenes, where $x=o+\ell\mathrm{d}\ (\ell>0)$ are sampled points along a ray originating at $o$ with direction $\mathrm{d}$ and $\mathbf{t}$ denotes timestamp. Then, they leverage volumetric rendering to render images,
\begin{align}
    C=\sum_{i}\omega_{i} c_{i},\ \text{where}\ \omega_{i}=e^{-\sum_{j<i}\tau_{j}(\ell_{j+1}-\ell_{j})}(1-e^{\tau_{i}(\ell_{i+1}-\ell_{i})}). \label{eq:volumetric_render}
\end{align}

Following 4D-fy \cite{fridovich2023k} and iNGP \cite{muller2022instant}, we use one multi-resolution spatial grid $\mathbf{P}_{xyz}$ and one spatial-time planes $\mathbf{P}_{xyzt}$ as 4D representation. Here, both $\mathbf{P}_{xyz}$ and $\mathbf{P}_{xyzt}$ use hash tables to store learnable features. Then, we acquire spatial-time features ${\mathbf{f}}$ through interpolation and hash lookup on $\mathbf{P}_{xyz}$ and $\mathbf{P}_{xyzt}$. Finally, ${\mathbf{f}}$ are decoded into density and view-independent color using tiny MLPs,
\begin{align}
    \psi:{\mathbf{f}}\mapsto \tau,\ \ \phi:{\mathbf{f}}\mapsto c.
\end{align}
The entire set of trainable parameters is denoted as $\theta_{\text{4D}}$. We can optimize our dynamic NeRF by using the multi-view videos generated from 4DM, however, 4DM can only produce four orthogonal viewpoints at one time. Training with such sparse views often results in overfitting to the training viewpoints, as presented in Fig.~\ref{fig:direct_on_videos}. To mitigate this, we leverage 4D-aware SDS to optimize the dynamic NeRF, enabling effective rendering from novel viewpoints across the temporal dimension, which is crucial for 4D generation.

\textbf{4D-aware SDS.} Once 4DM is trained, we employ 4D-aware SDS loss to guide the optimization of our 4D representation. To be concrete, we utilize Eqn.~\ref{eq:volumetric_render} to render four $F$ frames video $V_{r}$ with timestamps $\mathbf{t}=\{\mathbf{t_{1}}, \mathbf{t_{2}},...,\mathbf{t_{F}}\}$ from four orthogonal viewpoints $\mathbf{c}_{\text{mv}}$. Our 4D-aware SDS injects Gaussian noise $\epsilon$ into $V_{r}$ at a diffusion timestep $t$ and passes to our multi-view video diffusion to provide gradients to update $\theta_{\text{4D}}$,
\begin{align}
    \nabla_{\theta_{\text{4D}}} \cal{L}_{\text{4D-SDS}}\approx \mathbb{E}_{(\mathbf{c}_{\text{mv}}, \mathbf{t},\epsilon,\mathit{t})}\bigg [\text{2}(\textit{V}_{\textit{r}}-\hat{\textit{V}_{\text{0}}})\frac{\partial {\textit{V}_{\textit{r}}}}{\partial {\theta_{\text{4D}}}}\bigg ],
\end{align}
\noindent where $\hat{\textit{V}_{\text{0}}}$ denotes the pseudo ground truth denoised from 4DM with the input video $V$ and viewpoints $\mathbf{c}_{\text{mv}}$ as condition. Here, we replace original $\epsilon$-based SDS loss with $x_{0}$-reconstruction loss as in \cite{shi2023mvdream}.

\textbf{Anchor Loss.} Accurately estimating the elevation and azimuth of input monocular video within the canonical coordination is challenging, making it difficult to use the input video directly as supervision signals. Therefore, we utilize 4DM to produce four orthogonal videos conditioned on the input video and select the one with the viewpoint closest to that of the input video as the anchor video. This approach ensures that the anchor video maintains the same quality as the input and improves the results. Moreover, 4DM is currently limited to generating multi-view videos with 8 frames. When the input video exceeds 8 frames, we must apply our multi-view video diffusion model multiple times to generate anchor videos. However, this process may lead to temporally inconsistent results due to the stochasticity of the diffusion model, particularly when the viewpoint is far from the input video as shown in Fig.~\ref{fig:multi-4d}. This inconsistency would degrade the 4D generation performance. Finally, we devise an anchor loss $\cal{L}_{\text{a}}$ based on the anchor video to enhance the appearance details and facilitate the learning of dynamic NeRF. Since it is challenging for 4DM to ensure pixel-to-pixel alignment of the anchor video, we follow \cite{chen2024v3d} to use image-level perceptual loss, \ie LPIPS \cite{zhang2018unreasonable} and SSIM, for dynamic NeRF optimization, 
\begin{equation}
    \cal{L}_{\text{a}}=\lambda_{\text{1}}\cal\text{LPIPS}(\mathit{I_{r}}, \mathit{I_{a}}) + \lambda_{\text{2}}\cal\text{D-SSIM}(\mathit{I_{r}}, \mathit{I_{a}}),
\end{equation}
where $\mathit{I_{r}}$ and $\mathit{I_{a}}$ represent the rendered video and anchor video, $\lambda$ is the loss weight. Consequently, our total loss function for 4D generation is,
\begin{align}
    \cal{L}_{\text{4D}}=\cal{L}_{\text{4D-SDS}}+\cal{L}_{\text{a}}+\lambda_{\text{3}}\cal{L}_{\text{orient}}+\lambda_{\text{4}}\cal{L}_{\text{opacity}} + \lambda_{\text{5}}\cal{L}_{\text{sparse}},
\end{align}
\noindent here $\cal{L}_{\text{orient}}$, $\cal{L}_{\text{opacity}}$, and $\cal{L}_{\text{sparse}}$ are regularization loss in DreamFusion \cite{poole2022dreamfusion}.

\begin{figure}[t]
    \centering
    %\small\includegraphics[width=0.8\linewidth]{assets/teaser.pdf}
    \includegraphics[width=\linewidth]{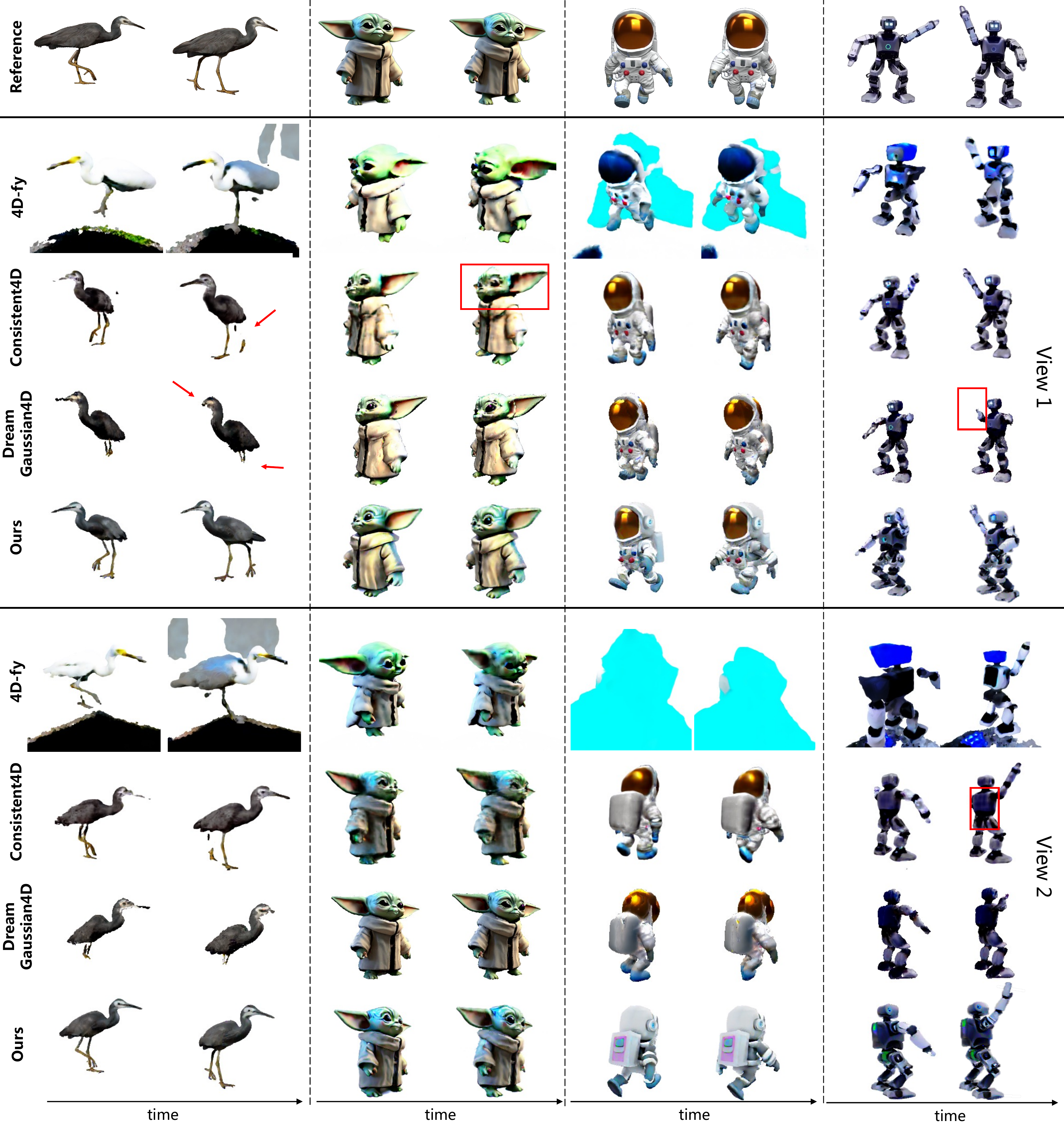}
    \vspace{-1.5em}
    \caption{4D generation comparisons with 4D-fy \cite{bahmani20234d}, Consistent4D \cite{jiang2023consistent4d}, and DreamGaussian4D \cite{ren2023dreamgaussian4d}.}
    % \vspace{-1em}
    \label{fig:4d_generation}
\end{figure}

\section{Experiments}

\noindent\textbf{Implementation Details.} We implement 4DM under the Stable Diffusion framework and initialize it from the checkpoint of ImageDream. We train 4DM with multi-view videos with 256$\times$256 resolutions for 30,000 steps with a batch size of 32, using the AdamW optimizer with a learning rate of 1e-4. The training takes about 2 days with 16 NVIDIA Tesla A100 GPUs. Additionally, for 4D generation experiments, we optimize dynamic NeRF representation in an end-to-end manner, avoiding utilizing multiple stages as in previous works.

\noindent\textbf{Baselines.} To evaluate our method, we compare to two video-to-4D approaches, namely Consistent4D \cite{jiang2023consistent4d} and DreamGaussian4D \cite{ren2023dreamgaussian4d}, and one text-to-4D approach 4D-fy \cite{bahmani20234d}. We extend 4D-fy to video-prompt 4D generation by using ImageDream as the 3D-aware diffusion model. 4D-fy introduces hybrid SDS to blend gradients from multiple pre-trained diffusion models to create 4D contents. Consistent4D is the first study focusing on the video-to-4D task. They utilize a 3D-aware diffusion model to optimize a cascade dynamic NeRF and propose a consistency loss to address spatial-temporal inconsistency. DreamGaussian4D leverages 4D Gaussian Splatting for faster training.

\subsection{Comparisons on 4D Generation}
\noindent\textbf{Qualitative Evaluation.} To validate 4Diffusion for 4D generation, we compare it to Consistent4D \cite{jiang2023consistent4d}, DreamGaussian4D \cite{ren2023dreamgaussian4d}, and 4D-fy \cite{bahmani20234d} on monocular video-to-4D task. Here, we use 3 real-world videos and 3 synthetic videos from the Consistent4D dataset, as well as 3 images from the ImageDream. As discussed in Sec.A.1 of the supplementary materials, for text-image pairs from ImageDream, we utilize SVD to generate input videos. We illustrate the results in Fig.~\ref{fig:4d_generation}. 4D-fy \cite{bahmani20234d} achieves state-of-the-art results on text-to-4D task and can be simply extended to video-prompt 4D generation by replacing 3D-aware diffusion model. Here, we utilize ImageDream as the 3D-aware diffusion model in 4D-fy. 4D-fy produces 4D contents with inconsistent temporal appearance, sometimes diverging significantly from the input video, as depicted in the first two columns of Fig.~\ref{fig:4d_generation}. This is primarily because integrating gradients from multiple diffusion models is difficult and they face challenges in multi-view spatial-temporal modeling. Consistent4D is the first work for 4D generation from monocular video. They employ an interpolation loss between two frames to enhance spatial-temporal consistency. However, they lack temporal consistency across frames, leading to poor appearance quality and flickers. DreamGaussian4D generates 4D contents with a blurred appearance and inaccurate geometry because GS struggles to model thin structures and large motions under unconstrained situations. In contrast, 4Diffusion generates high-quality 4D content with 4DM, which captures multi-view spatial-temporal correlations in a unified manner. Overall, our method achieves superior results, demonstrating its effectiveness. For more visualization results, please refer to our supplementary materials.

\label{sec:multi_quan}
\noindent\textbf{Quantitative Evaluation.} 
 We select 5 test cases from Objaverse, each consisting of a monocular input video and four orthogonal ground truth videos, which are not included in the training data, to evaluate our model. To evaluate image quality, we leverage CLIP-I \cite{radford2021learning} to measure the similarity. We also calculate FVD to evaluate the video quality. We compute LPIPS \cite{zhang2018unreasonable} and PSNR metrics to evaluate the spatial consistency. Here, we use ground truth videos for novel viewpoints to compute the above metrics. Moreover, we compute CLIP-C between frames in each synthetic video to evaluate temporal consistency. Tab.~\ref{tab:quan_4d} presents the results, clearly demonstrating that 4Diffusion outperforms other methods on all metrics. 

% \begin{table}[t]
% \centering
%     \caption{Quantitative evaluation on 4D generation.}
%     \label{tab:quan_4d}
%     % \resizebox{\textwidth}{!}{%
%     \begin{tabular}{c@{\quad}c@{\quad}c@{\quad}c}
%     \specialrule{.15em}{.1em}{.1em}
%     & \multicolumn{1}{c}{\text{Image quality}} & \text{Tem. Con.} & \text{Video quality}  \\ 
%     \cmidrule(r){2-2}\cmidrule(r){3-3}\cmidrule(r){4-4}
%     & CLIP-I$\uparrow$ & CLIP-C$\uparrow$ & FVD$\downarrow$ \\ 
%     \midrule
%     Consistent4D\cite{jiang2023consistent4d} & 0.8713 & 0.9641 & 1246.9  \\
%     DreamGaussian4D\cite{ren2023dreamgaussian4d} & 0.8587 & 0.9625 & 1451.2  \\
%     4D-fy\cite{bahmani20234d} & 0.8287 & 0.9379 & 2283.6  \\
%     \midrule
%     Ours(w/o $\cal{L}_{\text{4D-SDS}}$) & 0.7945 & 0.9423 & 2103.2 \\
%     Ours(w/o $\cal{L}_{\text{a}}$) & 0.8518 & 0.9557 & 1647.0 \\
%     Ours & \textbf{0.8803} & \textbf{0.9654} & \textbf{1196.8} \\
%     \specialrule{.15em}{.1em}{.1em}
%     \end{tabular}%
%     % }
% \end{table}

\begin{table}[t]
\centering
    \caption{Quantitative evaluation on 4D generation.}
    \label{tab:quan_4d}
    % \resizebox{\textwidth}{!}{%
    \begin{tabular}{c@{\quad}c@{\quad}c@{\quad}c@{\quad}c@{\quad}c}
    \specialrule{.15em}{.1em}{.1em}
    & \multicolumn{1}{c}{\text{Image quality}} & \text{Tem. Con.} & \text{Video quality} & \multicolumn{2}{c}{\text{Spa. Con.}}  \\ 
    \cmidrule(r){2-2}\cmidrule(r){3-3}\cmidrule(r){4-4}\cmidrule(r){5-6}
    & CLIP-I$\uparrow$ & CLIP-C$\uparrow$ & FVD$\downarrow$ &LPIPS$\downarrow$ & PSNR$\uparrow$ \\ 
    \midrule
    4D-fy\cite{bahmani20234d} & 0.8658 & 0.9487 & 1042.3 & 0.2254 & 14.24  \\
    Consistent4D\cite{jiang2023consistent4d} & 0.9216 & 0.9723 & 706.07 & 0.1593 & 16.70 \\
    DreamGaussian4D\cite{ren2023dreamgaussian4d} & 0.8898 & 0.9710 & 760.18 & 0.1793 & 15.97  \\
    \midrule
    Ours(w/o $\cal{L}_{\text{4D-SDS}}$) & 0.8195 & 0.9503 & 1546.4 & 0.2356 & 13.92 \\
    Ours(w/o $\cal{L}_{\text{a}}$) & 0.8823 & 0.9720 & 853.57 & 0.1589 &  17.20 \\
    Ours & \textbf{0.9310} & \textbf{0.9798} & \textbf{417.63} & \textbf{0.1199} & \textbf{19.07} \\
    \specialrule{.15em}{.1em}{.1em}
    \end{tabular}%
    % }
\end{table}

\begin{figure}[t]
    \centering
    %\small\includegraphics[width=0.8\linewidth]{assets/teaser.pdf}
    \includegraphics[width=\linewidth]{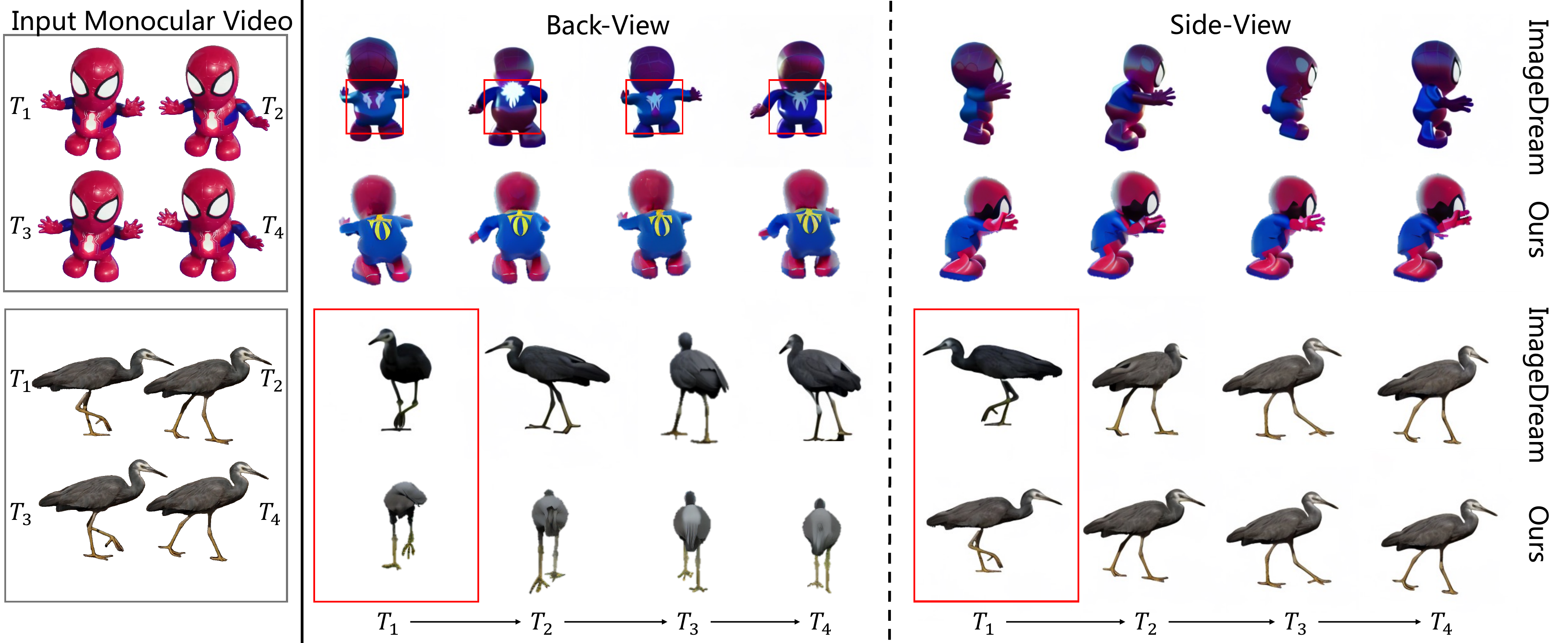}
    %\vspace{-1.5em}
    \caption{The illustration of synthesized multi-view videos from 4DM and ImageDream \cite{wang2023imagedream}. 4DM produces more spatial-temporal consistent results than ImageDream. $T$ denotes the timestep of video clips. All results are generated from DDIM \cite{song2020denoising} sampler.}
    % \vspace{-1em}
    \label{fig:multi-view-video}
\end{figure}

\subsection{Multi-view Video Generation}
\noindent\textbf{Qualitative Evaluation.} In this section, we evaluate the multi-view video generation quality produced by 4DM using the same input videos as described in qualitative evaluation in Sec~\ref{sec:multi_quan}. We employ ImageDream to synthesize a set of multi-view images as pseudo multi-view video by taking each frame of the input monocular video as an image prompt. Fig.~\ref{fig:multi-view-video} illustrates results with the first two columns corresponding to the input monocular video. Although ImageDream excels at synthesizing spatially consistent images, it struggles to model temporal correlations, leading to inconsistent temporal appearances, such as the icon on the back of Spiderman. Comparatively, 4DM effectively captures reasonable temporal information using the motion module, even when trained on a small curated dataset. Moreover, our model preserves the generalization ability of ImageDream, allowing us to generate high-fidelity multi-view videos, even beyond the distribution of our training dataset. As the last two rows of Fig.~\ref{fig:multi-view-video} show, 4DM produces spatially consistent videos by sharing information across spatial and temporal dimensions to constraint the generation process while ImageDream occasionally fails to generate videos coherent to the viewpoint.

\noindent\textbf{Quantitative Evaluation.} We use the same test cases described in quantitative evaluation in Sec~\ref{sec:multi_quan}, alongside the test data provided by Consistent4D, to evaluate 4DM. To account for the stochasticity of the diffusion model, we conduct five runs for each test case and report the average metrics. Tab.~\ref{tab:quan_multi_view_video} shows comparative results. Despite the comparable performance in CLIP-I, 4DM excels in generating spatial-temporally consistent multi-view videos, a primary focus of our research. This is evidenced by the superior performance on metrics such as CLIP-C, FVD, LPIPS, and PSNR, which better capture the spatial and temporal fidelity of video content. These metrics demonstrate that our method effectively balances image quality with temporal consistency, making it a robust solution for multi-view video generation.

\begin{table}[t]
\centering
    \caption{Quantitative evaluation on multi-view video generation. Here, we employ Consistent4D test dataset to evaluate 4DM and ImageDream. 'Spa. Con.' and 'Tem. Con.' refer to spatial consistency and temporal consistency, respectively.}
    \label{tab:quan_multi_view_video}
    % \resizebox{\textwidth}{!}{
    \begin{tabular}{cccccc} 
        \specialrule{.15em}{.1em}{.1em}
        & \multicolumn{1}{c}{\text{Image quality}} & \text{{\quad}Tem. Con. {\quad}} & \text{Video Quality} & \multicolumn{2}{c}{\text{{\quad}Spa. Con.{\quad}}}  \\ 
        \cmidrule(r){2-2}\cmidrule(r){3-3}\cmidrule(r){4-4}\cmidrule(r){5-6}
        & CLIP-I$\uparrow$ & CLIP-C$\uparrow$ & FVD$\downarrow$ & LPIPS$\downarrow$ & PSNR$\uparrow$ \\ 
        \midrule
        ImageDream\cite{wang2023imagedream} & 0.9165 & 0.9320 & 465.94  & 0.1536 & 16.57  \\
        Ours(w/ whole) & 0.8872 & 0.9478 & 583.79 & 0.1763 & 15.28  \\
        % \specialrule{.15em}{.1em}{.1em}
        Ours(4DM) & \textbf{0.9260} & \textbf{0.9601} & \textbf{427.34} & \textbf{0.1346} & \textbf{17.88} \\
        \specialrule{.15em}{.1em}{.1em}
    \end{tabular}
    % }
\end{table}

\subsection{Ablation study and analysis}

\begin{wrapfigure}{R}{0.5\textwidth}
\centering{
    %\small\includegraphics[width=0.8\linewidth]{assets/teaser.pdf}
    \includegraphics[width=\linewidth]{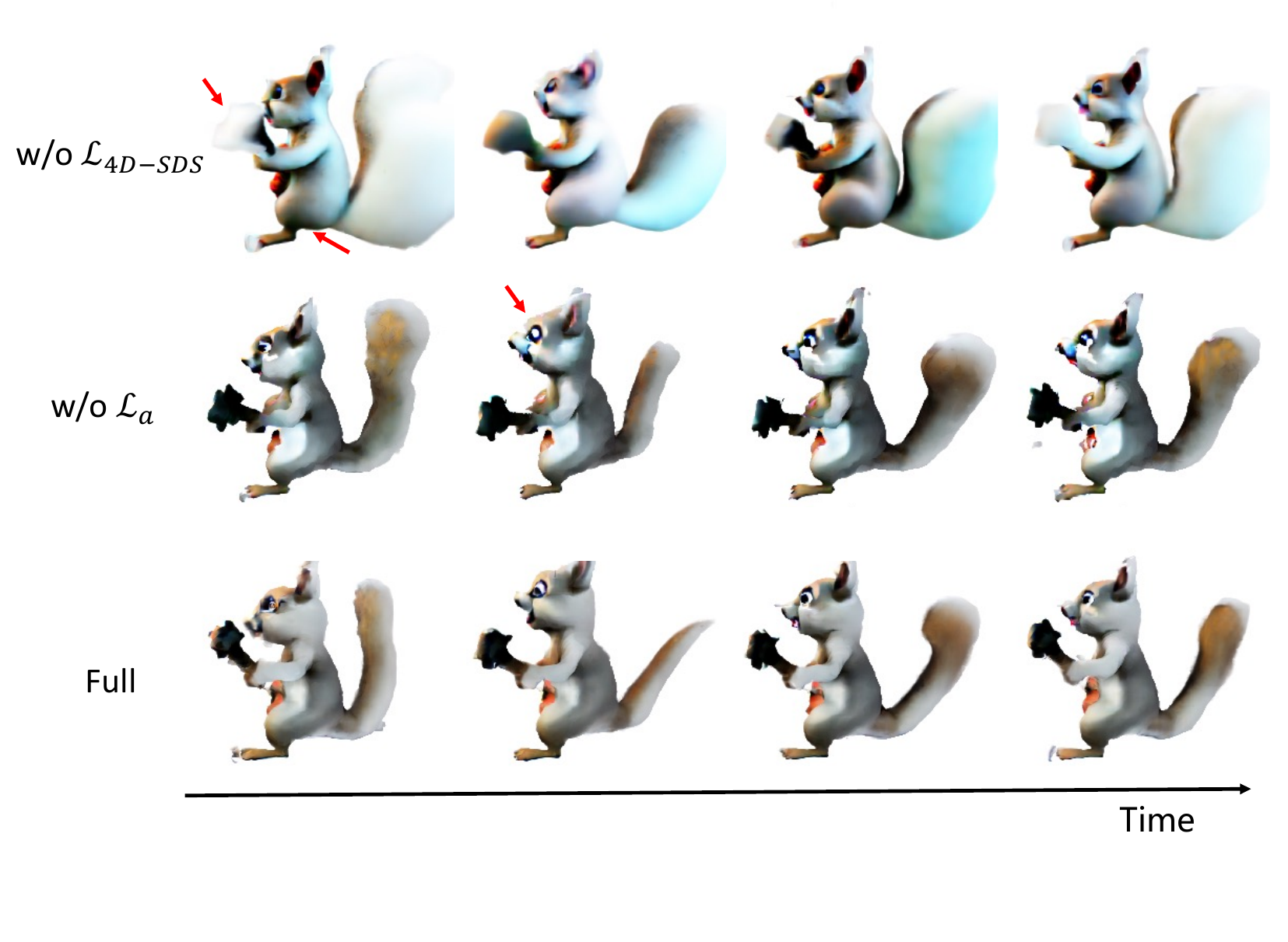}
    %\vspace{-1.5em}
    \vspace{-30pt}
    \caption{Ablation studies on 4D-aware SDS loss and the anchor loss.}
    \label{fig:ablation}
}
\end{wrapfigure}
\textbf{Effectiveness of the Curated Multi-view Video Dataset.} To evaluate the importance and effectiveness of the selected high-quality multi-view videos, we use the entire animated 3D shapes from Objaverse and render multi-view videos to fine-tune 4DM (Ours w/ whole). The results are shown in Tab.~\ref{tab:quan_multi_view_video}. Given the presence of numerous flawed data within the entire dataset, it compromises the image quality of ImageDream and encounters challenges in precisely capturing spatial-temporal correlations, demonstrating the importance of high-quality datasets for fine-tuning 4DM.

\noindent\textbf{4D-aware SDS Loss.} To evaluate the effect of our 4D-aware SDS loss, we substitute the 4DM with ImageDream and use 3D-aware SDS loss based on ImageDream to optimize dynamic NeRF representation. As Fig.~\ref{fig:ablation} depicted, inconsistent temporal textures, such as the leg of the squirrel, emerge due to the lack of temporal modeling of ImageDream, underscoring the significance of capturing spatial-temporal correlations in 4DM. The quantitative results presented in Tab.~\ref{tab:quan_4d} indicate the significance of our 4D-aware SDS loss.

\noindent\textbf{Anchor Loss.} We also assess the impact of the proposed anchor loss. As illustrated in Fig.~\ref{fig:ablation}, capturing detailed appearance features, such as the eyes of the squirrel, proves challenging without the anchor loss. Conversely, the anchor images furnish visual clues to facilitate the learning of 4D representation, resulting in high-quality 4D content. The quantitative results Tab.~\ref{tab:quan_4d} demonstrate the crucial role of our anchor loss.

\section{Conclusion}
In this paper, we present 4Diffusion for 4D generation from a monocular video. Our method proposes a multi-view video diffusion model 4DM based on a 3D-aware diffusion model for multi-view video generation and provides multi-view spatial-temporal guidance for 4D generation. 4DM captures spatial-temporal correlations and preserves the characteristics of the origin 3D-aware diffusion model even when training on a small curated dataset. Then, we combine 4D-aware SDS loss and an anchor loss based on 4DM to optimize our hash-encoded dynamic NeRF, resulting in spatial-temporally consistent 4D contents coherent with the input monocular video.

\clearpage  % TODO REVIEW/FINAL: This \clearpage needs to be removed from both review and camera-ready versions.

% \section*{References}

\section*{Acknowlegements}
The work is supported by the National Key R\&D Program of China (No. 2022ZD0160102), the National Natural Science Foundation of China under Grant No. 62102150, and the Science and Technology Commission of Shanghai Municipality under Grant No. 23QD1400800.

\bibliographystyle{plain}
\bibliography{ref}

\begin{thebibliography}{10}

\bibitem{deepfloyd}
Deepfloyd. \url{https://github.com/deep-floyd/IF}.
\newblock 2023.

\bibitem{zeroscope}
Zeroscope text-to-video model. \url{https://huggingface. co/cerspense/zeroscope_v2_576w}.
\newblock 2023.

\bibitem{anciukevivcius2023renderdiffusion}
Titas Anciukevi{\v{c}}ius, Zexiang Xu, Matthew Fisher, Paul Henderson, Hakan Bilen, Niloy~J Mitra, and Paul Guerrero.
\newblock Renderdiffusion: Image diffusion for 3d reconstruction, inpainting and generation.
\newblock In {\em CVPR}, pages 12608--12618, 2023.

\bibitem{bahmani20234d}
Sherwin Bahmani, Ivan Skorokhodov, Victor Rong, Gordon Wetzstein, Leonidas Guibas, Peter Wonka, Sergey Tulyakov, Jeong~Joon Park, Andrea Tagliasacchi, and David~B Lindell.
\newblock 4d-fy: Text-to-4d generation using hybrid score distillation sampling.
\newblock In {\em CVPR}, pages 7996--8006, 2024.

\bibitem{blattmann2023stable}
Andreas Blattmann, Tim Dockhorn, Sumith Kulal, Daniel Mendelevitch, Maciej Kilian, Dominik Lorenz, Yam Levi, Zion English, Vikram Voleti, Adam Letts, Varun Jampani, and Rodin Rombach.
\newblock Stable video diffusion: Scaling latent video diffusion models to large datasets.
\newblock {\em arXiv preprint arXiv:2311.15127}, 2023.

\bibitem{cao2023hexplane}
Ang Cao and Justin Johnson.
\newblock Hexplane: A fast representation for dynamic scenes.
\newblock In {\em CVPR}, pages 130--141, 2023.

\bibitem{caron2021emerging}
Mathilde Caron, Hugo Touvron, Ishan Misra, Herv{\'e} J{\'e}gou, Julien Mairal, Piotr Bojanowski, and Armand Joulin.
\newblock Emerging properties in self-supervised vision transformers.
\newblock In {\em ICCV}, pages 9650--9660, 2021.

\bibitem{chan2022efficient}
Eric~R Chan, Connor~Z Lin, Matthew~A Chan, Koki Nagano, Boxiao Pan, Shalini De~Mello, Orazio Gallo, Leonidas~J Guibas, Jonathan Tremblay, Sameh Khamis, Tero Karras, and Gordon Wetzstein.
\newblock Efficient geometry-aware 3d generative adversarial networks.
\newblock In {\em CVPR}, pages 16123--16133, 2022.

\bibitem{chen2023fantasia3d}
Rui Chen, Yongwei Chen, Ningxin Jiao, and Kui Jia.
\newblock Fantasia3d: Disentangling geometry and appearance for high-quality text-to-3d content creation.
\newblock In {\em ICCV}, pages 22246--22256, 2023.

\bibitem{chen2024v3d}
Zilong Chen, Yikai Wang, Feng Wang, Zhengyi Wang, and Huaping Liu.
\newblock V3d: Video diffusion models are effective 3d generators.
\newblock {\em arXiv preprint arXiv:2403.06738}, 2024.

\bibitem{deitke2023objaverse}
Matt Deitke, Dustin Schwenk, Jordi Salvador, Luca Weihs, Oscar Michel, Eli VanderBilt, Ludwig Schmidt, Kiana Ehsani, Aniruddha Kembhavi, and Ali Farhadi.
\newblock Objaverse: A universe of annotated 3d objects.
\newblock In {\em CVPR}, pages 13142--13153, 2023.

\bibitem{erkocc2023hyperdiffusion}
Ziya Erko{\c{c}}, Fangchang Ma, Qi~Shan, Matthias Nie{\ss}ner, and Angela Dai.
\newblock Hyperdiffusion: Generating implicit neural fields with weight-space diffusion.
\newblock In {\em ICCV}, pages 14300--14310, 2023.

\bibitem{fridovich2023k}
Sara Fridovich-Keil, Giacomo Meanti, Frederik~Rahb{\ae}k Warburg, Benjamin Recht, and Angjoo Kanazawa.
\newblock K-planes: Explicit radiance fields in space, time, and appearance.
\newblock In {\em CVPR}, pages 12479--12488, 2023.

\bibitem{guo2023i2v}
Xun Guo, Mingwu Zheng, Liang Hou, Yuan Gao, Yufan Deng, Pengfei Wan, Di~Zhang, Yufan Liu, Weiming Hu, Zhengjun Zha, Haibin Huang, and Chongyang Ma.
\newblock I2v-adapter: A general image-to-video adapter for video diffusion models.
\newblock {\em arXiv preprint arXiv:2312.16693}, 2023.

\bibitem{liuthreestudio}
Yuan-Chen Guo, Ying-Tian Liu, Ruizhi Shao, Christian Laforte, Vikram Voleti, Guan Luo, Chia-Hao Chen, Zi-Xin Zou, Chen Wang, Yan-Pei Cao, and Song-Hai Zhang.
\newblock threestudio: A unified framework for 3d content generation.
\newblock \url{https://github.com/threestudio-project/threestudio}, 2023.

\bibitem{guo2023animatediff}
Yuwei Guo, Ceyuan Yang, Anyi Rao, Yaohui Wang, Yu~Qiao, Dahua Lin, and Bo~Dai.
\newblock Animatediff: Animate your personalized text-to-image diffusion models without specific tuning.
\newblock In {\em ICLR}, 2024.

\bibitem{gupta20233dgen}
Anchit Gupta, Wenhan Xiong, Yixin Nie, Ian Jones, and Barlas O{\u{g}}uz.
\newblock 3dgen: Triplane latent diffusion for textured mesh generation.
\newblock {\em arXiv preprint arXiv:2303.05371}, 2023.

\bibitem{hong2023lrm}
Yicong Hong, Kai Zhang, Jiuxiang Gu, Sai Bi, Yang Zhou, Difan Liu, Feng Liu, Kalyan Sunkavalli, Trung Bui, and Hao Tan.
\newblock Lrm: Large reconstruction model for single image to 3d.
\newblock In {\em ICLR}, 2024.

\bibitem{huang2023sc}
Yi-Hua Huang, Yang-Tian Sun, Ziyi Yang, Xiaoyang Lyu, Yan-Pei Cao, and Xiaojuan Qi.
\newblock Sc-gs: Sparse-controlled gaussian splatting for editable dynamic scenes.
\newblock In {\em CVPR}, pages 4220--4230, 2024.

\bibitem{jiang2023consistent4d}
Yanqin Jiang, Li~Zhang, Jin Gao, Weimin Hu, and Yao Yao.
\newblock Consistent4d: Consistent 360° dynamic object generation from monocular video.
\newblock In {\em ICLR}, 2024.

\bibitem{kerbl20233d}
Bernhard Kerbl, Georgios Kopanas, Thomas Leimk{\"u}hler, and George Drettakis.
\newblock 3d gaussian splatting for real-time radiance field rendering.
\newblock {\em TOG}, 42(4), 2023.

\bibitem{kingma2013auto}
Diederik~P Kingma and Max Welling.
\newblock Auto-encoding variational bayes.
\newblock In {\em ICLR}, 2014.

\bibitem{li2023instant3d}
Jiahao Li, Hao Tan, Kai Zhang, Zexiang Xu, Fujun Luan, Yinghao Xu, Yicong Hong, Kalyan Sunkavalli, Greg Shakhnarovich, and Sai Bi.
\newblock Instant3d: Fast text-to-3d with sparse-view generation and large reconstruction model.
\newblock In {\em ICLR}, 2024.

\bibitem{li2023nerfacc}
Ruilong Li, Hang Gao, Matthew Tancik, and Angjoo Kanazawa.
\newblock Nerfacc: Efficient sampling accelerates nerfs.
\newblock In {\em ICCV}, pages 18537--18546, 2023.

\bibitem{li2022neural}
Tianye Li, Mira Slavcheva, Michael Zollhoefer, Simon Green, Christoph Lassner, Changil Kim, Tanner Schmidt, Steven Lovegrove, Michael Goesele, Richard Newcombe, and Zhaoyang Lv.
\newblock Neural 3d video synthesis from multi-view video.
\newblock In {\em CVPR}, pages 5521--5531, 2022.

\bibitem{lin2023magic3d}
Chen-Hsuan Lin, Jun Gao, Luming Tang, Towaki Takikawa, Xiaohui Zeng, Xun Huang, Karsten Kreis, Sanja Fidler, Ming-Yu Liu, and Tsung-Yi Lin.
\newblock Magic3d: High-resolution text-to-3d content creation.
\newblock In {\em CVPR}, pages 300--309, 2023.

\bibitem{lin2023consistent123}
Yukang Lin, Haonan Han, Chaoqun Gong, Zunnan Xu, Yachao Zhang, and Xiu Li.
\newblock Consistent123: One image to highly consistent 3d asset using case-aware diffusion priors.
\newblock {\em arXiv preprint arXiv:2309.17261}, 2023.

\bibitem{ling2023align}
Huan Ling, Seung~Wook Kim, Antonio Torralba, Sanja Fidler, and Karsten Kreis.
\newblock Align your gaussians: Text-to-4d with dynamic 3d gaussians and composed diffusion models.
\newblock In {\em CVPR}, pages 8576--8588, 2024.

\bibitem{liu2023zero}
Ruoshi Liu, Rundi Wu, Basile Van~Hoorick, Pavel Tokmakov, Sergey Zakharov, and Carl Vondrick.
\newblock Zero-1-to-3: Zero-shot one image to 3d object.
\newblock In {\em ICCV}, pages 9298--9309, 2023.

\bibitem{liu2023humangaussian}
Xian Liu, Xiaohang Zhan, Jiaxiang Tang, Ying Shan, Gang Zeng, Dahua Lin, Xihui Liu, and Ziwei Liu.
\newblock Humangaussian: Text-driven 3d human generation with gaussian splatting.
\newblock In {\em CVPR}, pages 6646--6657, 2024.

\bibitem{liu2023syncdreamer}
Yuan Liu, Cheng Lin, Zijiao Zeng, Xiaoxiao Long, Lingjie Liu, Taku Komura, and Wenping Wang.
\newblock Syncdreamer: Generating multiview-consistent images from a single-view image.
\newblock In {\em ICLR}, 2024.

\bibitem{long2023wonder3d}
Xiaoxiao Long, Yuan-Chen Guo, Cheng Lin, Yuan Liu, Zhiyang Dou, Lingjie Liu, Yuexin Ma, Song-Hai Zhang, Marc Habermann, Christian Theobalt, and wenping Wang.
\newblock Wonder3d: Single image to 3d using cross-domain diffusion.
\newblock In {\em CVPR}, pages 9970--9980, 2024.

\bibitem{luo2023scalable}
Tiange Luo, Chris Rockwell, Honglak Lee, and Justin Johnson.
\newblock Scalable 3d captioning with pretrained models.
\newblock In {\em NeurIPS}, 2023.

\bibitem{melas2023realfusion}
Luke Melas-Kyriazi, Iro Laina, Christian Rupprecht, and Andrea Vedaldi.
\newblock Realfusion: 360° reconstruction of any object from a single image.
\newblock In {\em CVPR}, pages 8446--8455, 2023.

\bibitem{mildenhall2021nerf}
Ben Mildenhall, Pratul~P Srinivasan, Matthew Tancik, Jonathan~T Barron, Ravi Ramamoorthi, and Ren Ng.
\newblock Nerf: Representing scenes as neural radiance fields for view synthesis.
\newblock In {\em ECCV}, 2020.

\bibitem{muller2023diffrf}
Norman M{\"u}ller, Yawar Siddiqui, Lorenzo Porzi, Samuel~Rota Bulo, Peter Kontschieder, and Matthias Nie{\ss}ner.
\newblock Diffrf: Rendering-guided 3d radiance field diffusion.
\newblock In {\em CVPR}, pages 4328--4338, 2023.

\bibitem{muller2022instant}
Thomas M{\"u}ller, Alex Evans, Christoph Schied, and Alexander Keller.
\newblock Instant neural graphics primitives with a multiresolution hash encoding.
\newblock {\em TOG}, 41(4):1--15, 2022.

\bibitem{poole2022dreamfusion}
Ben Poole, Ajay Jain, Jonathan~T Barron, and Ben Mildenhall.
\newblock Dreamfusion: Text-to-3d using 2d diffusion.
\newblock In {\em ICLR}, 2023.

\bibitem{qian2023magic123}
Guocheng Qian, Jinjie Mai, Abdullah Hamdi, Jian Ren, Aliaksandr Siarohin, Bing Li, Hsin-Ying Lee, Ivan Skorokhodov, Peter Wonka, Sergey Tulyakov, and Bernard Ghanem.
\newblock Magic123: One image to high-quality 3d object generation using both 2d and 3d diffusion priors.
\newblock In {\em ICLR}, 2024.

\bibitem{radford2021learning}
Alec Radford, Jong~Wook Kim, Chris Hallacy, Aditya Ramesh, Gabriel Goh, Sandhini Agarwal, Girish Sastry, Amanda Askell, Pamela Mishkin, Jack Clark, Gretchen Krueger, and Sutskever Ilya.
\newblock Learning transferable visual models from natural language supervision.
\newblock In {\em ICML}, pages 8748--8763, 2021.

\bibitem{raj2023dreambooth3d}
Amit Raj, Srinivas Kaza, Ben Poole, Michael Niemeyer, Nataniel Ruiz, Ben Mildenhall, Shiran Zada, Kfir Aberman, Michael Rubinstein, Jonathan Barron, Yuanzhen Li, and Varun Jampani.
\newblock Dreambooth3d: Subject-driven text-to-3d generation.
\newblock In {\em ICCV}, 2023.

\bibitem{ren2023dreamgaussian4d}
Jiawei Ren, Liang Pan, Jiaxiang Tang, Chi Zhang, Ang Cao, Gang Zeng, and Ziwei Liu.
\newblock Dreamgaussian4d: Generative 4d gaussian splatting.
\newblock {\em arXiv preprint arXiv:2312.17142}, 2023.

\bibitem{rombach2022high}
Robin Rombach, Andreas Blattmann, Dominik Lorenz, Patrick Esser, and Bj{\"o}rn Ommer.
\newblock High-resolution image synthesis with latent diffusion models.
\newblock In {\em CVPR}, pages 10684--10695, 2022.

\bibitem{shi2023mvdream}
Yichun Shi, Peng Wang, Jianglong Ye, Mai Long, Kejie Li, and Xiao Yang.
\newblock Mvdream: Multi-view diffusion for 3d generation.
\newblock In {\em ICLR}, 2024.

\bibitem{shue20233d}
J~Ryan Shue, Eric~Ryan Chan, Ryan Po, Zachary Ankner, Jiajun Wu, and Gordon Wetzstein.
\newblock 3d neural field generation using triplane diffusion.
\newblock In {\em CVPR}, pages 20875--20886, 2023.

\bibitem{singer2022make}
Uriel Singer, Adam Polyak, Thomas Hayes, Xi~Yin, Jie An, Songyang Zhang, Qiyuan Hu, Harry Yang, Oron Ashual, Oran Gafni, Devi Parikh, Sonal Gupta, and Yaniv Taigman.
\newblock Make-a-video: Text-to-video generation without text-video data.
\newblock In {\em ICLR}, 2023.

\bibitem{singer2023text}
Uriel Singer, Shelly Sheynin, Adam Polyak, Oron Ashual, Iurii Makarov, Filippos Kokkinos, Naman Goyal, Andrea Vedaldi, Devi Parikh, Justin Johnson, and Yaniv Taigman.
\newblock Text-to-4d dynamic scene generation.
\newblock In {\em ICML}, pages 31915--31929, 2023.

\bibitem{song2020denoising}
Jiaming Song, Chenlin Meng, and Stefano Ermon.
\newblock Denoising diffusion implicit models.
\newblock In {\em ICLR}, 2021.

\bibitem{tang2024lgm}
Jiaxiang Tang, Zhaoxi Chen, Xiaokang Chen, Tengfei Wang, Gang Zeng, and Ziwei Liu.
\newblock Lgm: Large multi-view gaussian model for high-resolution 3d content creation.
\newblock In {\em ECCV}, pages 1--18, 2024.

\bibitem{tang2023dreamgaussian}
Jiaxiang Tang, Jiawei Ren, Hang Zhou, Ziwei Liu, and Gang Zeng.
\newblock Dreamgaussian: Generative gaussian splatting for efficient 3d content creation.
\newblock In {\em ICLR}, 2024.

\bibitem{tang2023make}
Junshu Tang, Tengfei Wang, Bo~Zhang, Ting Zhang, Ran Yi, Lizhuang Ma, and Dong Chen.
\newblock Make-it-3d: High-fidelity 3d creation from a single image with diffusion prior.
\newblock In {\em ICCV}, 2023.

\bibitem{wang2023imagedream}
Peng Wang and Yichun Shi.
\newblock Imagedream: Image-prompt multi-view diffusion for 3d generation.
\newblock {\em arXiv preprint arXiv:2312.02201}, 2023.

\bibitem{wang2023rodin}
Tengfei Wang, Bo~Zhang, Ting Zhang, Shuyang Gu, Jianmin Bao, Tadas Baltrusaitis, Jingjing Shen, Dong Chen, Fang Wen, Qifeng Chen, and Baining Guo.
\newblock Rodin: A generative model for sculpting 3d digital avatars using diffusion.
\newblock In {\em CVPR}, pages 4563--4573, 2023.

\bibitem{wang2023videofactory}
Wenjing Wang, Huan Yang, Zixi Tuo, Huiguo He, Junchen Zhu, Jianlong Fu, and Jiaying Liu.
\newblock Videofactory: Swap attention in spatiotemporal diffusions for text-to-video generation.
\newblock {\em arXiv preprint arXiv:2305.10874}, 2023.

\bibitem{wang2024prolificdreamer}
Zhengyi Wang, Cheng Lu, Yikai Wang, Fan Bao, Chongxuan Li, Hang Su, and Jun Zhu.
\newblock Prolificdreamer: High-fidelity and diverse text-to-3d generation with variational score distillation.
\newblock In {\em NeurIPS}, 2023.

\bibitem{wu20234d}
Guanjun Wu, Taoran Yi, Jiemin Fang, Lingxi Xie, Xiaopeng Zhang, Wei Wei, Wenyu Liu, Qi~Tian, and Xinggang Wang.
\newblock 4d gaussian splatting for real-time dynamic scene rendering.
\newblock In {\em CVPR}, pages 20310--20320, 2024.

\bibitem{wu2024sc4d}
Zijie Wu, Chaohui Yu, Yanqin Jiang, Chenjie Cao, Fan Wang, and Xiang Bai.
\newblock Sc4d: Sparse-controlled video-to-4d generation and motion transfer.
\newblock {\em arXiv preprint arXiv:2404.03736}, 2024.

\bibitem{xu2023dmv3d}
Yinghao Xu, Hao Tan, Fujun Luan, Sai Bi, Peng Wang, Jiahao Li, Zifan Shi, Kalyan Sunkavalli, Gordon Wetzstein, Zexiang Xu, et~al.
\newblock Dmv3d: Denoising multi-view diffusion using 3d large reconstruction model.
\newblock In {\em ICLR}, 2024.

\bibitem{yang2023consistnet}
Jiayu Yang, Ziang Cheng, Yunfei Duan, Pan Ji, and Hongdong Li.
\newblock Consistnet: Enforcing 3d consistency for multi-view images diffusion.
\newblock In {\em CVPR}, pages 7079--7088, 2024.

\bibitem{yariv2023mosaic}
Lior Yariv, Omri Puny, Natalia Neverova, Oran Gafni, and Yaron Lipman.
\newblock Mosaic-sdf for 3d generative models.
\newblock In {\em CVPR}, pages 4630--4639, 2024.

\bibitem{ye2023ip}
Hu~Ye, Jun Zhang, Sibo Liu, Xiao Han, and Wei Yang.
\newblock Ip-adapter: Text compatible image prompt adapter for text-to-image diffusion models.
\newblock {\em arXiv preprint arXiv:2308.06721}, 2023.

\bibitem{yin20234dgen}
Yuyang Yin, Dejia Xu, Zhangyang Wang, Yao Zhao, and Yunchao Wei.
\newblock 4dgen: Grounded 4d content generation with spatial-temporal consistency.
\newblock {\em arXiv preprint arXiv:2312.17225}, 2023.

\bibitem{zeng2024stag4d}
Yifei Zeng, Yanqin Jiang, Siyu Zhu, Yuanxun Lu, Youtian Lin, Hao Zhu, Weiming Hu, Xun Cao, and Yao Yao.
\newblock Stag4d: Spatial-temporal anchored generative 4d gaussians.
\newblock {\em arXiv preprint arXiv:2403.14939}, 2024.

\bibitem{zhang2018unreasonable}
Richard Zhang, Phillip Isola, Alexei~A Efros, Eli Shechtman, and Oliver Wang.
\newblock The unreasonable effectiveness of deep features as a perceptual metric.
\newblock In {\em CVPR}, pages 586--595, 2018.

\bibitem{zhao2023animate124}
Yuyang Zhao, Zhiwen Yan, Enze Xie, Lanqing Hong, Zhenguo Li, and Gim~Hee Lee.
\newblock Animate124: Animating one image to 4d dynamic scene.
\newblock {\em arXiv preprint arXiv:2311.14603}, 2023.

\bibitem{zheng2023unified}
Yufeng Zheng, Xueting Li, Koki Nagano, Sifei Liu, Otmar Hilliges, and Shalini De~Mello.
\newblock A unified approach for text-and image-guided 4d scene generation.
\newblock In {\em CVPR}, pages 7300--7309, 2024.

\bibitem{zou2023triplane}
Zi-Xin Zou, Zhipeng Yu, Yuan-Chen Guo, Yangguang Li, Ding Liang, Yan-Pei Cao, and Song-Hai Zhang.
\newblock Triplane meets gaussian splatting: Fast and generalizable single-view 3d reconstruction with transformers.
\newblock In {\em CVPR}, pages 10324--10335, 20234.

\end{thebibliography}

\clearpage  % TODO REVIEW/FINAL: This \clearpage needs to be removed from both review and camera-ready versions.

\appendix

\section{Supplemental material}

\subsection{More Implementation Details}

\begin{table}[htb]
\centering
    \caption{Hash encoding parameters of $P_{xyz}$ and $P_{xyzt}$}
    \label{tab:parameters}
    \begin{tabular}{c@{\quad}c}
    \specialrule{.15em}{.1em}{.1em}
    Parameter & Value  \\ 
    \midrule
    Number of levels & 16 \\
    Hash table size & $2^{19}$ \\
    Number of feature dimensions per level & 2 \\
    Coarsest resolution & 16 \\
    Finest resolution & 4096 \\
    \specialrule{.15em}{.1em}{.1em}
    \end{tabular}%
\end{table}

\begin{figure}[htb]
    \centering
    %\small\includegraphics[width=0.8\linewidth]{assets/dataset.pdf}
    \includegraphics[width=\linewidth]{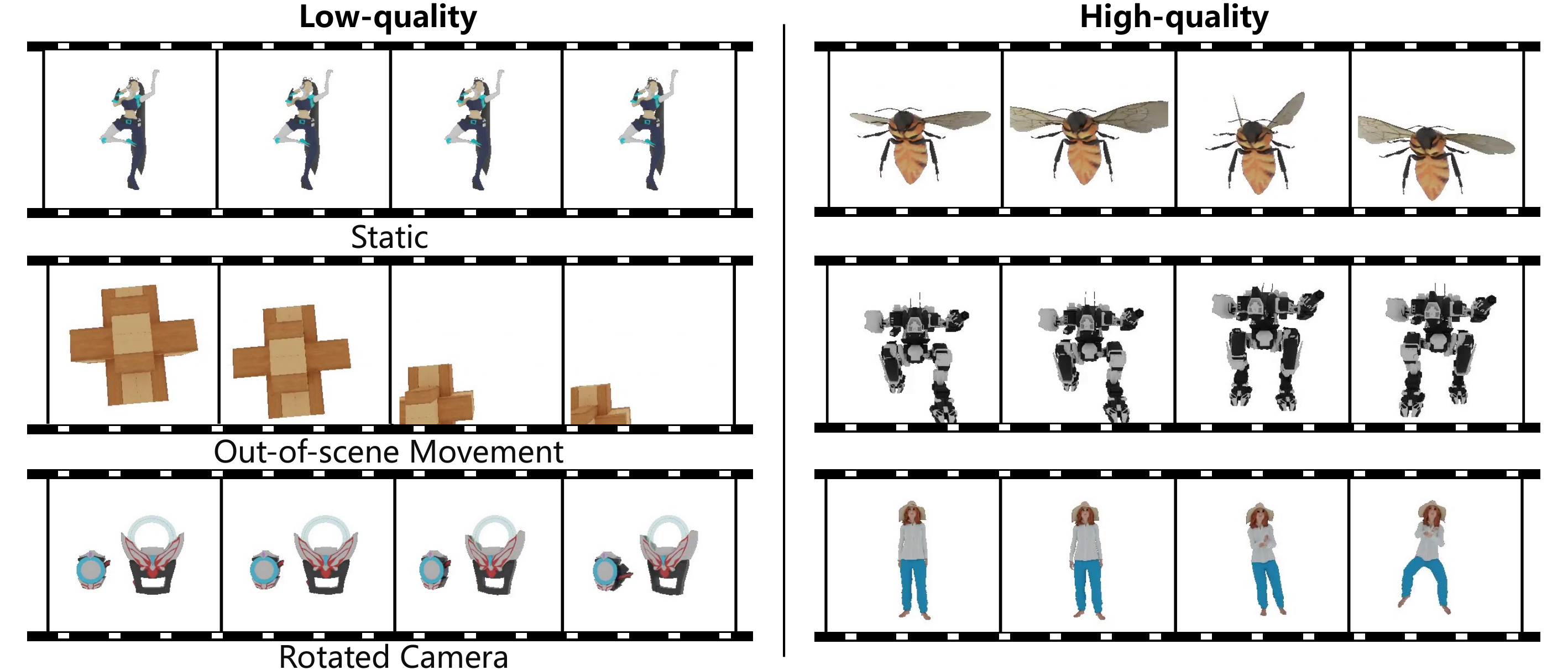}
    \vspace{-1.5em}
    \caption{The illustration of our training dataset. We manually filter out animated 3D data with static motion, out-of-scene movement, or rotated camera to curate a dataset with high-quality appearance and realistic motion. }
    % \vspace{-3em}
    \label{fig:dataset}
\end{figure}

\textbf{Datasets.} We utilize Objaverse dataset \cite{deitke2023objaverse} to train our multi-view diffusion model, as described in Sec.~\ref{sec:video_diffusion}. Objaverse dataset comprises a vast collection of 3D shapes with descriptive captions, tags, and animations. We manually filter out animated 3D shapes that contain static objects, out-of-scene movement, rotated cameras, or meaningless objects, resulting in 926 high-quality 3D animated models, as depicted in Fig.~\ref{fig:dataset}. We apply Blender to render 32 videos with azimuth angles uniformly ranging from $[-180^{\circ}, 180^{\circ}]$ and an elevation angle of $0^{\circ}$ for each animated 3D model. In our experiments, we use the dataset released by Consistent4D \cite{jiang2023consistent4d}, test cases from Objaverse, and text-image pairs from ImageDream \cite{wang2023imagedream} project page. Specifically, for text-image pairs, we leverage Stable Video Diffusion V1.1 \cite{blattmann2023stable} to produce monocular videos for 4D generation. 

\textbf{4D Generation.} We implement our 4D generation model under threestudio framework \cite{liuthreestudio}. Our hash-encoded dynamic NeRF representation utilizes the parameters detailed in Tab.~\ref{tab:parameters}. Following \cite{wang2024prolificdreamer, bahmani20234d}, we anneal the timesteps of diffusion models from $t\in [0.98, 0.98]$ to $t\in [0.02, 0.25]$ over the initial 5,000 iterations and set the diffusion CFG to 5.0. The loss weights $\lambda_{1}$, $\lambda_{2}$, $\lambda_{4}$ are set to 200, 100, and 100, respectively. Addtionally, $\lambda_{3}$ linearly increases from 10 to 1000 during the first 5,000 iterations and $\lambda_{5}$ is fixed at 100 after the initial 10,000 iterations. The model is trained with AdamW optimizer for 35,000 iterations with a learning rate of 1e-2 except for the decoded MLPs, where the learning rate is adjusted to 1e-3. It takes around 12 hours to train the model on one NVIDIA Tesla A100 GPU.

\textbf{Volume Rendering.} We employ NerfAcc \cite{li2023nerfacc} as our rendering pipeline, which leverages an occupancy grid to store the opacity of a scene. This approach accelerates volume rendering and reduces computations. We adopt a shared occupancy grid by representing the maximum opacity of the scene across all frames, facilitating its application to dynamic scenes. Additionally, we set the background of the rendered images to white. For the resolution of rendered images, we follow the configuration of ImageDream \cite{wang2023imagedream}. We maintain fixed camera distances at 1.1 to enhance the stability of the optimization process.

\subsection{More Results}

\begin{figure}
    \centering
    %\small\includegraphics[width=0.8\linewidth]{assets/dataset.pdf}
    \includegraphics[width=\linewidth]{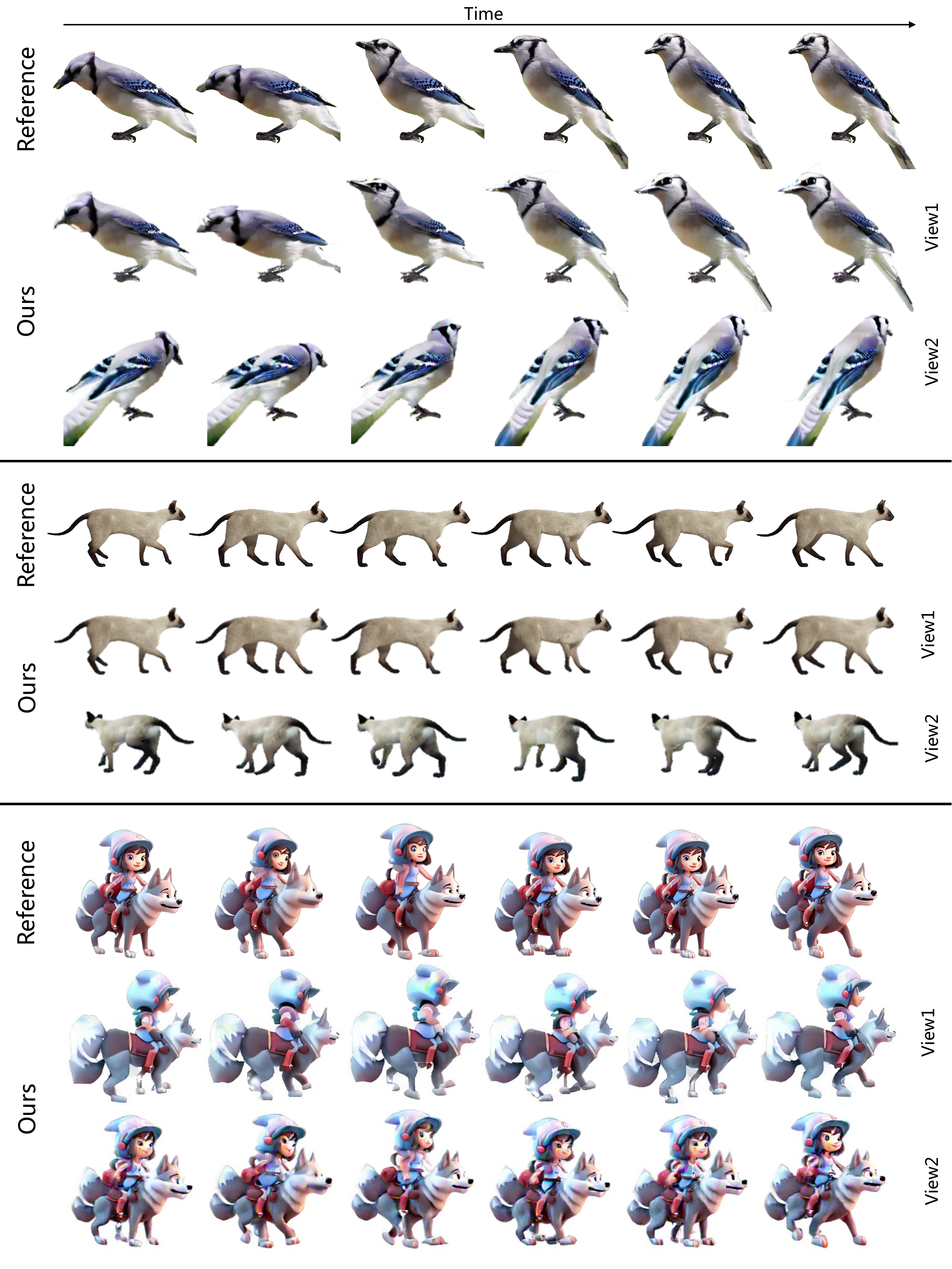}
    \caption{The illustration of 4D generation results of 4Diffusion.}
    \label{fig:supp_4d}
\end{figure}

\begin{figure}
    \centering
    %\small\includegraphics[width=0.8\linewidth]{assets/dataset.pdf}
    \includegraphics[width=\linewidth]{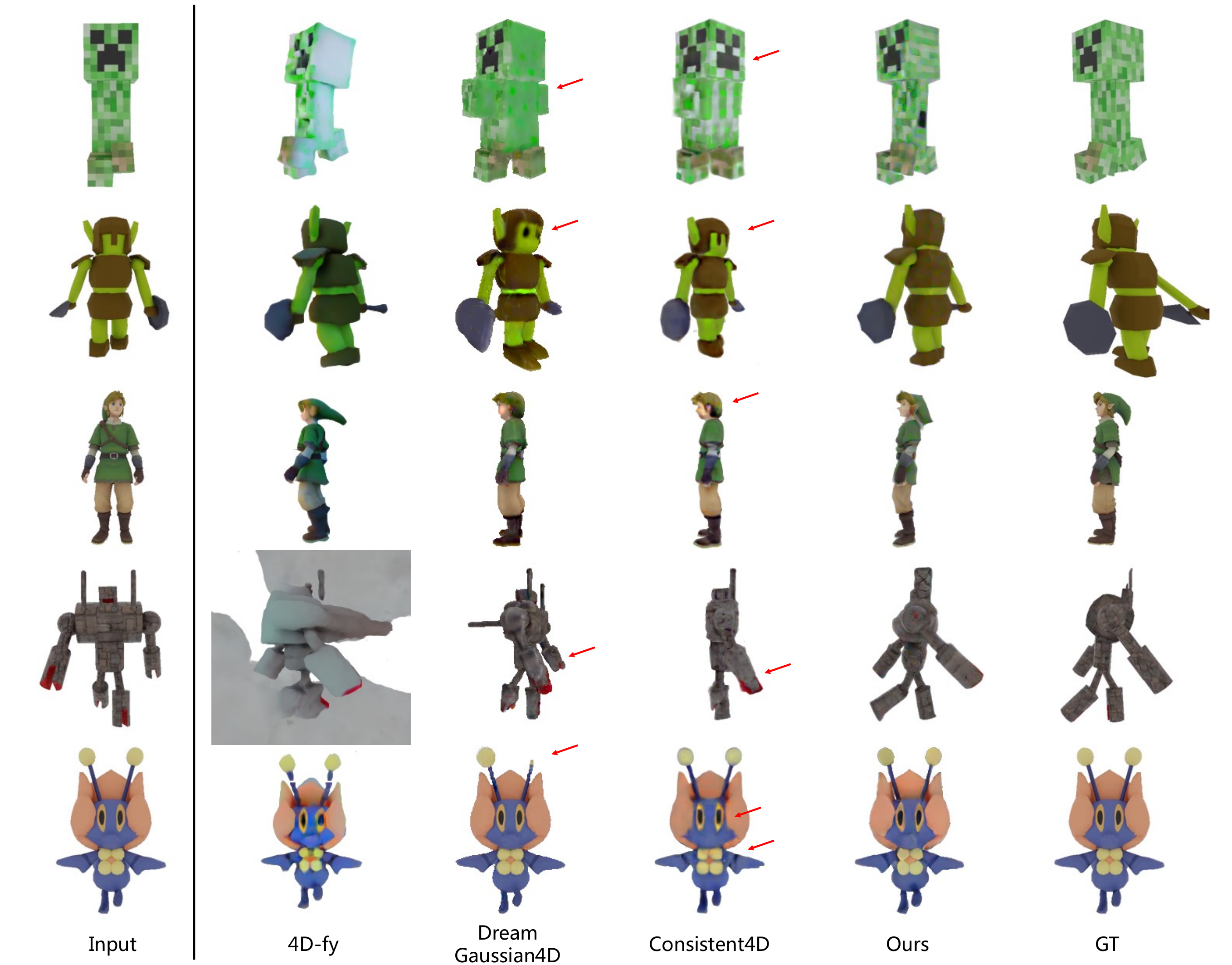}
    \caption{4D generation comparisons with 4D-fy, DreamGaussian4D, Consistent4D. These test cases are selected from Objaverse dataset, which are not included in the training data of 4DM.}
    \label{fig:rebuttal_4d}
\end{figure}
\textbf{4D Generation.} In Fig.~\ref{fig:supp_4d}, we showcase additional results of our 4D generation results. To gain a more intuitive understanding, we encourage readers to view the supplementary videos. Moreover, we show more comparisons on 4D generation as shown in Fig.~\ref{fig:rebuttal_4d}, Consistent4D and DreamGaussian4D encounter the multi-face problem while our method generates spatial-temporally consistent contents. 

\begin{figure}
    \centering
    %\small\includegraphics[width=0.8\linewidth]{assets/dataset.pdf}
    \includegraphics[width=\linewidth]{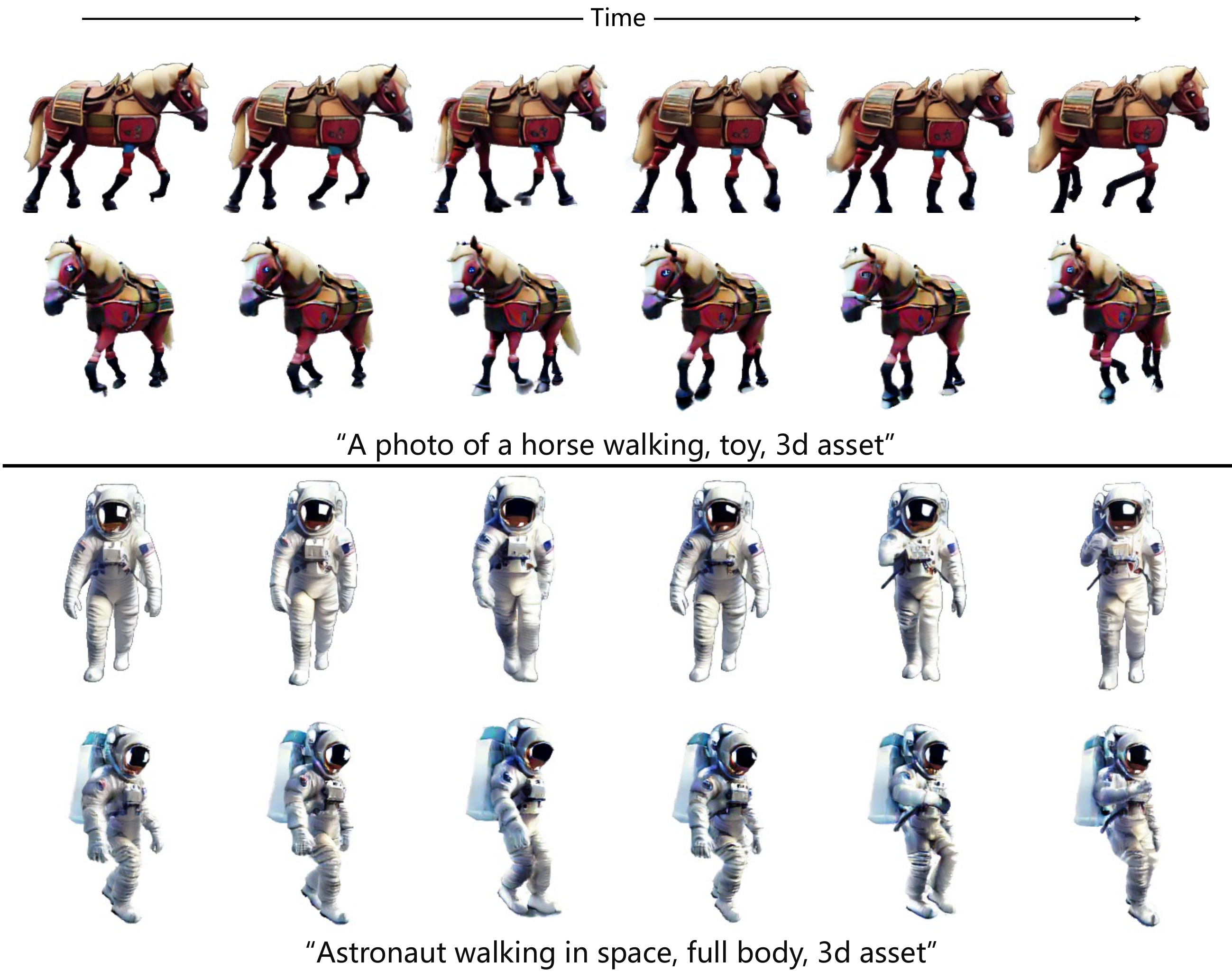}
    \caption{The illustration of text-to-4D results of 4Diffusion.}
    \label{fig:supp_4d_2}
\end{figure}

\textbf{Text-to-4D.} To Further validate the effectiveness of 4Diffusion, we conduct experiments on text-to-4D task. Specifically, we first employ SDXL to generate images conditioned on text prompts. Subsequently, we utilize SVD V1.1 to produce monocular videos for 4D generation. Finally, we follow the procedure outlined in the main paper to generate 4D content from the monocular videos. As illustrated in Figure \ref{fig:supp_4d_2}, our approach yields high-quality 4D content from text prompts, thereby demonstrating its effectiveness.

\subsection{Limitations} 
Our method can be improved in the following aspects: \textbf{1)} Our multi-view video diffusion model is constrained by the capability of the base model and the scale of high-quality training data. We believe that improving the base model and scaling up the high-quality dataset can obtain a better model. \textbf{2)} Our 4D generation pipeline relies on heavily volumetric rendering, causing slow training speed. We believe advances in 3D and GS can potentially solve these problems.

\subsection{Broader Impacts} 
Our work paves the way for high-quality 4D content generation, reducing the extensive manual effort for artists and novices. Although our method is not designed for generating humans, it may be extended and misused, potentially influencing human perceptions. 

%%%%%%%%%%%%%%%%%%%%%%%%%%%%%%%%%%%%%%%%%%%%%%%%%%%%%%%%%%%%

\end{document}